\documentclass[10pt]{article} % For LaTeX2e
\usepackage[accepted]{tmlr_files/tmlr}
\usepackage{amsmath,amsfonts,bm}

\def\eqref#1{equation~\ref{#1}}
\def\1{\bm{1}}

\DeclareMathAlphabet{\mathsfit}{\encodingdefault}{\sfdefault}{m}{sl}
\SetMathAlphabet{\mathsfit}{bold}{\encodingdefault}{\sfdefault}{bx}{n}

\DeclareMathOperator*{\argmax}{arg\,max}

\definecolor{orange}{rgb}{1,0.5,0}
\definecolor{gr}{rgb}{0,0.65,0}
\definecolor{mygray}{gray}{0.95}

\newcommand{\x}{\mathbf{x}}

\newcommand{\w}{\mathbf{w}}
\newcommand{\cl}[1]{\mathbf{\boldsymbol{\phi}}_{#1}}

\newcommand{\karate}{\emph{karate}}
\newcommand{\polbooks}{\emph{polbooks}}
\newcommand{\football}{\emph{football}}
\newcommand{\polblogs}{\emph{polblogs}}

\newcommand{\animal}{\emph{animal}}
\newcommand{\group}{\emph{group}}

\newcommand{\mammal}{\emph{mammal}}

\newcommand{\occupation}{\emph{occupation}}
\newcommand{\rodent}{\emph{rodent}}

\newcommand{\an}{\emph{an}}
\newcommand{\oms}{\emph{oms}}
\newcommand{\pt}{\emph{pt}}

\newcommand{\comment}[1]{}

\newcommand{\xmark}{\ding{55}}%
\newcommand{\cmark}{\ding{51}}%

\usepackage{hyperref}
\usepackage{url}

\usepackage{xcolor}
\usepackage{dsfont}
\usepackage{amssymb}
\usepackage{bm}
\usepackage{algpseudocode}
\usepackage{amsmath}
\usepackage{multirow}
\usepackage{pifont}% http://ctan.org/pkg/pifont
\usepackage{colortbl}
\usepackage{booktabs}
\usepackage{adjustbox}
\usepackage{makecell}
\usepackage{subfig}

\DeclareMathOperator{\arcosh}{arcosh}

\title{Hyperbolic Random Forests}

\author{\name Lars Doorenbos \email lars.doorenbos@unibe.ch \\
      University of Bern
      \AND
      \name Pablo Márquez-Neila \email pablo.marquez@unibe.ch \\
      University of Bern
      \AND
      \name Raphael Sznitman \email raphael.sznitman@unibe.ch\\
      University of Bern
      \AND
      \name Pascal Mettes \email p.s.m.mettes@uva.nl\\
      University of Amsterdam}

\def\month{05}  % Insert correct month for camera-ready version
\def\year{2024} % Insert correct year for camera-ready version
\def\openreview{\url{https://openreview.net/forum?id=pjKcIzvXWR}} % Insert correct link to OpenReview for camera-ready version

\begin{document}

\maketitle

\begin{abstract}
Hyperbolic space is becoming a popular choice for representing data due to the hierarchical structure \textemdash~whether implicit or explicit \textemdash~of many real-world datasets. Along with it comes a need for algorithms capable of solving fundamental tasks, such as classification, in hyperbolic space.
Recently, multiple papers have investigated hyperbolic alternatives to hyperplane-based classifiers, such as logistic regression and SVMs. While effective, these approaches struggle with more complex hierarchical data. We, therefore, propose to generalize the well-known random forests to hyperbolic space.
We do this by redefining the notion of a split using horospheres. Since finding the globally optimal split is computationally intractable, we find candidate horospheres through a large-margin classifier. To make hyperbolic random forests work on multi-class data and imbalanced experiments, we furthermore outline new methods for combining classes based on the lowest common ancestor and class-balanced large-margin losses. Experiments on standard and new benchmarks show that our approach outperforms both conventional random forest algorithms and recent hyperbolic classifiers.
\end{abstract}

\section{Introduction}

Machine learning in hyperbolic space is gaining more and more attention, and hyperbolic representations of data have already found success in numerous domains, such as natural language processing~\citep{nickel2017poincare,tai2022hyperbolic} computer vision~\citep{ahmad2022fisheyehdk,khrulkov2020hyperbolic,ghadimi2021hyperbolic}, graphs \citep{chami2019hyperbolic,liu2019hyperbolic,sun2021hyperbolic}, recommender systems \citep{sun2021hgcf}, and more. Hyperbolic space is a natural choice for data with a hierarchical structure due to the fact that the available space grows exponentially when moving away from the origin. Therefore, it can be seen as a continuous version of a graph-theoretical tree \citep{nickel2018learning}.

With the rise of datasets embedded in hyperbolic space comes a need for algorithms that can successfully operate on them~\citep{cho2019large}.
As a result, many methods specifically designed for hyperbolic space have been proposed that tackle a variety of machine learning tasks, such as clustering~\citep{monath2019gradient}, regression~\citep{marconi2020hyperbolic}, and classification~\citep{ganea2018nns,chien2021highly,cho2019large,pan2023provably,weber2020robust}. 

\begin{figure}[t]
  \centering
  \setlength\tabcolsep{2pt}
  \begin{tabular}{ccc}
    \includegraphics[width=0.275\linewidth]{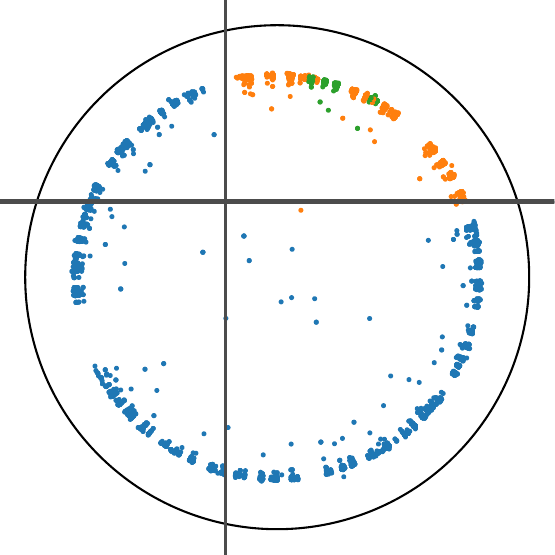} &
    \includegraphics[width=0.275\linewidth]{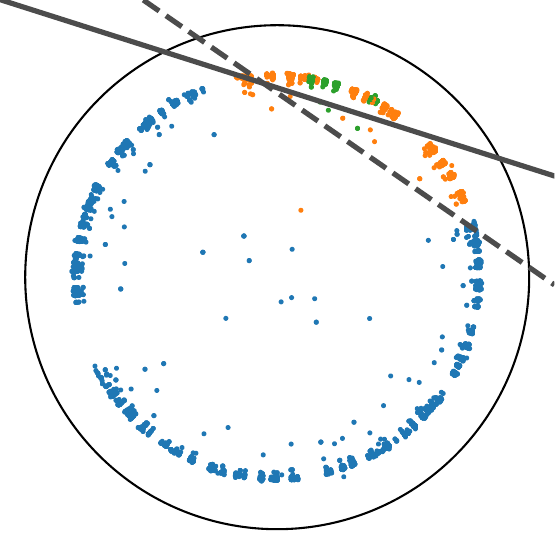} &
    \includegraphics[width=0.275\linewidth]{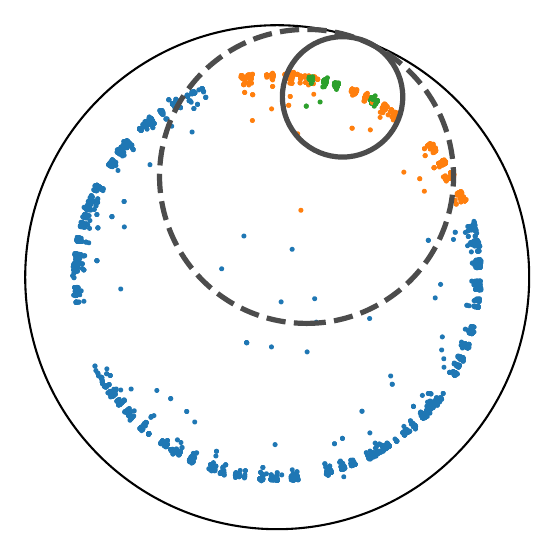} \\
    (a) & (b) & (c) \\
  \end{tabular}
    \caption{\textbf{Motivation for hyperbolic random forests.} Two splits of (a) an axis-aligned Euclidean, (b) an oblique Euclidean, and (b) a hyperbolic decision tree of depth two. The data is a continuous embedding of a tree split into three nested classes. The inductive biases of linear decision boundaries are inappropriate for efficiently capturing the underlying geometry, and more splits are needed to be effective. 
    }
    \label{fig:teaser}
\end{figure}

% \begin{figure}[t]
%   \centering
%   \setlength\tabcolsep{2pt}
%   \begin{tabular}{cc}
%     \includegraphics[width=0.48\linewidth]{images/teaser/euc_teaser.pdf} &
%     \includegraphics[width=0.48\linewidth]{images/teaser/hyp_teaser.pdf} \\
%     (a) & (b) \\
%   \end{tabular}
%     \caption{\textbf{Motivation for hyperbolic random forests.} The two splits of (a) an oblique Euclidean and (b) a hyperbolic decision tree of depth two with the highest information gain. The data is a tree embedded in hyperbolic space, split into three nested classes. The inductive biases of linear decision boundaries are not appropriate for the underlying geometry. 
%     }
%     \label{fig:teaser}
% \end{figure}

Current hyperbolic classification algorithms, such as hyperbolic support vector machines~\citep{fan2023horocycle} or hyperbolic logistic regression~\citep{ganea2018nns}, have shown promising results, but still struggle with complex datasets. In Euclidean space, there is a long history of success of tree-based random forest algorithms in this regard, and they remain among the most popular learning algorithms for many data types~\citep{caruana2006empirical,fernandez2014we}. We argue that decision trees and, by extension, random forests are well-suited for hyperbolic space due to their shared hierarchical structure.
%However, to our knowledge, no hyperbolic variant of random forests exists to date \LD{do/can we keep this sentence given the preprints online}. We strive to fill this gap.
% \ld{Tree-based algorithms remain among the most popular learning algorithms for many data types. Despite their hierarchical nature, to date, no tree-based algorithms exist for hyperbolic space. Our work is a first step towards further research in this direction. }
Figure~\ref{fig:teaser} illustrates the need for a classifier that fits the underlying geometry and the benefits of a hyperbolic tree-based classifier. 
Euclidean hyperplane splits (Figure~\ref{fig:teaser}(a/b)) are ineffective at capturing the structure of the data. In contrast, our hyperbolic splits (Figure~\ref{fig:teaser}(c)) are more appropriate, where the nested splits clearly show how the smallest green class is a subclass of the larger orange one.

Tree-based algorithms are built by recursively applying a splitting function; hence, to generalize the concept of a tree-based classifier to hyperbolic space, we need to define a hyperbolic splitting function. For this purpose, we propose to use \textit{horospheres}, which share several desirable properties with the hyperplanes used in Euclidean trees. Due to a combinatorial explosion, finding the optimal horosphere by enumeration is computationally intractable. We propose to tackle data splitting as a binary horosphere classification task. To deal with multiple classes in a hierarchically consistent manner, we extend our splitting operator using hyperclasses. We show how to obtain hyperclasses through clustering in hyperbolic space with the lowest common ancestor. We also outline a class-balancing large-margin loss for dealing with long-tailed data.

We extensively evaluate our approach on canonical benchmarks from previous works on hyperbolic classification. Additionally, we introduce three new multi-class classification experiments. We show that our method is superior to both competing hyperbolic classifiers as well as their Euclidean counterparts in these settings. 
Summarized, our contributions are (1)~a generalization of random forests to hyperbolic space using horospheres, dubbed HoroRF, (2) two extensions to enable effective learning in multi-class and imbalanced settings, and (3)~a thorough evaluation of HoroRF, showing its advantage over other methods, both Euclidean and hyperbolic.
\section{Related Work}

\subsection{Hyperbolic machine learning}
Machine learning in hyperbolic space has gained traction due to its inherent hierarchical and compact nature. Foundational work showed that hyperbolic space is superior to Euclidean space for embedding hierarchies in a continuous space, allowing for embeddings with minimal distortion \citep{nickel2017poincare,ganea2018hyperbolic}. Empowered by these results, learning with hyperbolic embeddings has recently been successfully used for various problems. \citet{chami2019hyperbolic} and \citet{liu2019hyperbolic} showed how to generalize graph networks to hyperbolic space, while \citet{tifrea2018poincar} showed the potential of hyperbolic word embeddings. In the visual domain, hyperbolic embeddings have been shown to improve image segmentation \citep{atigh2022hyperbolic}, zero-shot recognition \citep{liu2020hyperbolic}, image-text representation learning \citep{desai2023hyperbolic}, and more. Hyperbolic embeddings have also been effective for biology \citep{klimovskaia2020poincare} and in recommender systems \citep{sun2021hgcf}. We refer to recent surveys for a more complete overview \citep{peng2021hyperbolic,yang2022hyperbolic,mettes2023hyperbolic}.

For classification specifically, various traditional classifiers such as logistic regression~\citep{ganea2018nns}, neural networks~\citep{ganea2018nns,shimizu2021hyperbolic,van2023poincar}, and support vector machines (SVM)~\citep{cho2019large,fan2023horocycle,fang2021kernel,weber2020robust} now have counterparts in hyperbolic space. In this work, we strive to go beyond single hyperplane/gyroplane decision boundaries per class and bring random forests to hyperbolic space. 

Our method is built on the concept of horospheres, which are the level sets of the Busemann function to ideal points. Ideal points have recently succeeded in multiple areas, such as supervised classification~\citep{ghadimi2021hyperbolic} and self-supervised learning~\citep{durrant2023hmsn}. Horospheres, in particular, have found applications in dimensionality reduction~\citep{chami2021horopca}, as well as generalizing SVMs~\citep{fan2023horocycle} and neural networks~\citep{sonoda2022fully} to hyperbolic space, but we are the first to use them as building blocks for random forests. Concurrent with our work, \citet{chlenski2023fast} also propose a hyperbolic random forest model where they rely on geodesics rather than horospheres and work in the hyperboloid model of hyperbolic space instead of the Poincar\'{e} disk model.

\subsection{Random Forests}
Random Forests remain a popular choice of classifier to this day due to their high performance, speed, and insensitivity to hyperparameters~\citep{grinsztajn2022tree}. In their original paper~\citep{breiman2001random}, two versions of random forests were proposed: one with axis-aligned splits based on a single feature and one with oblique splits based on linear combinations of features. While the optimal axis-aligned split can be found by exhaustive search, finding the optimal oblique split is NP-complete~\citep{heath1993induction}. As a result, many works have developed heuristics to find good oblique splits in a reasonable time.

One line of work makes use of meta-heuristics such as hill climbing~\citep{murthy1993oc1}, simulated annealing~\citep{heath1993induction}, or genetic algorithms~\citep{cantu2003inducing}. Other approaches train one or more binary classifiers at every node, choosing among the resulting hyperplanes to split the data. Examples include using linear discriminant analysis, ridge regression~\citep{menze2011oblique}, or support vector machines~\citep{do2010classifying}. For multi-class cases, typically, heuristics are employed to partition them into two hyperclasses~\citep{katuwal2018enhancing}. We follow the binary classifier approach for our hyperbolic random forests and introduce a hyperclass heuristic specifically designed for hyperbolic space based on hierarchical clustering~\cite{chami2020trees}. 

\section{Hyperbolic Random Forests}

\subsection{The Poincar\'e ball model}
We follow the convention from previous works using horospheres~\citep{chami2021horopca,fan2023horocycle} and hyperbolic machine learning more broadly~\citep{ganea2018nns,khrulkov2020hyperbolic,chien2021highly,shimizu2021hyperbolic} and make use of the Poincar\'e ball model of hyperbolic space.
The Poincar\'e ball model is defined by the metric space $(\mathds{B}^n_c, g^\mathds{B}_c)$ for a given negative curvature $c$, which we set to 1 throughout this work, where 
\begin{equation}
    \mathds{B}^n_1 = \{\x \in \mathds{R}^n : \|\x\| < 1\}
\end{equation}
with $\|\cdot\|$ the standard Euclidean norm, and 
\begin{equation}
    g^\mathds{B}_1(\x) = \frac{2}{1 - \|\x\|^2}\mathbf{I}_n.
\end{equation}
From here on, we will omit the curvature subscript for clarity. The distance between two points is given by
\begin{equation}
\label{eq:dist}
    d_\mathds{B}(\mathbf{a}, \mathbf{b}) = \arcosh\left(1 + 2\frac{\|\mathbf{a} - \mathbf{b}\|^2}{(1 - \|\mathbf{a}\|^2)(1 - \|\mathbf{b}\|^2)}\right),
\end{equation}
which is the length of the geodesic arc connecting them.
Extending geodesics to infinity in one direction leads to a point on the boundary of the Poincar\'e ball. These points on the boundary are known as \textit{ideal points}. The set of all ideal points thus lies on the hypersphere $\mathds{S}^{n-1}$, and they can be interpreted as directions in hyperbolic space~\citep{chami2021horopca}.

\subsection{HoroRF}

To generalize random forests to hyperbolic space, we require a way to split data points recursively for any number of classes, regardless of the class frequency distribution, into two partitions with low impurity. In Euclidean space, this splitting function outputs hyperplanes to split the data. For hyperbolic random forests, we propose to use horospheres.

\subsubsection{Formalization}

Horospheres are the level sets of the Busemann function~\citep{busemann2012geometry}, which calculates the normalized distance to infinity in a given direction. It can be expressed in closed form in the Poincar\'e model:
\begin{equation}
    B_\w(\x) = \log\frac{\|\w - \x\|^2}{1 - \|\x\|^2}.
\end{equation}
As a result, horospheres $\pi_{\w,b}$ are parameterized by an ideal point $\w\in\mathds{S}^{n-1}$, and a distance to that point $b\in\mathds{R}$, which is the radius of the horosphere. 
Our goal is to find the optimal horosphere $\pi_{\w,b}^\prime$ that minimizes the impurity of the resulting partitions or, equivalently, maximizes the information gain. Consider a node $S$ with data $\{(\x_i, y_i)\}_{i=1}^N$ with $\x\in\mathds{B}^n$ and $y\in Y$. The optimal horosphere is given by 
\begin{equation}
\label{eq:ig}
    \pi_{\w,b}^\prime = \argmax_{\pi_{\w,b}}(H(S) - I_{in} - I_{out}),
\end{equation}
where $H$ is an impurity measure, and $I_{in}$ and $I_{out}$ give the impurity of the set of points inside and outside of the horosphere, respectively:
\begin{align}
\label{eq:imp}
    I_{in} &= \frac{N_{in}}{N} H(\{\x \in S | B_\w(\x) < b\}), \nonumber\\
    I_{out} &= \frac{N_{out}}{N}H(\{\x \in S | B_\w(\x) \geq b\}). 
\end{align}
Here, $N_{in}$ and $N_{out}$ denote the number of samples inside and outside the horosphere.
Similar to oblique decision trees in Euclidean space, finding the globally optimal horosphere is computationally infeasible.
As such, we need to find approximate solutions.

\subsubsection{Finding approximate solutions}

As is conventional in (oblique) decision tree literature (\emph{e.g.}, \citet{do2010classifying,menze2011oblique,katuwal2020heterogeneous}), we employ a binary classifier at every node to find candidate solutions. We need a hyperbolic classifier that will allow us to find splits that fit the underlying geometry. Large margin classifiers based on horospheres guarantee a globally optimal solution, bypassing hyperbolic hyperplane-based methods that fail to converge~\citep{weber2020robust} or are limited to two dimensions~\citep{chien2021highly}. Moreover, the state-of-the-art hyperbolic SVM, HoroSVM, is based on horospheres and outperforms hyperbolic hyperplane-based methods~\citep{fan2023horocycle}.
For this reason, we build upon HoroSVM.
%HoroSVM uses horospheres to create a large-margin classifier for hyperbolic space.
For a labeled dataset $D = \{(\x_i, y_i)\}_{i=1}^N$ with $\x_i\in\mathds{B}^n$ and $y_i\in\{-1, 1\}$, it optimizes the convex loss function
\begin{equation}
\label{eq:svmloss}
    \boldsymbol\ell(\mu,\w,o;D) = \frac{1}{2}\mu^2 + C\sum^{N}_{i=1}\max\left(0, 1 - y_i(\mu B_\w^{-1}(\x_i) - o)\right),
\end{equation}
where
\begin{equation}
    B_\w^{-1}(\x) = \log\frac{1 - \|\x\|^2}{\|\w-\x\|^2},
\end{equation}
with $\mu,o \in \mathds{R}^+$ and $\w\in\mathds{S}^{n-1}$. The slack hyperparameter $C$ controls the trade-off between misclassification tolerance and margin size. Solving for HoroSVM thus results in the three parameters $\mu, \w$, and $o$. In order to transform these results into a horosphere $\pi_{\w,b}$, we set $b = -\frac{o}{\mu}$ and use $\w$ directly. 

We repeat the classification $K$ times in a one-versus-rest setting, with $K$ the number of classes, and compute the information gain for all resulting horospheres. We select the horosphere with the highest information gain to split the data. We refer to this splitting procedure as the \texttt{HoroSplitter}. While this base version of the \texttt{HoroSplitter} already achieves competitive results, it naturally has limited capacity to deal with imbalanced data, and the one-versus-rest set-up limits its effectiveness in multi-class settings. Therefore, we introduce two additional components to the \texttt{HoroSplitter} that make it more appropriate specifically as a building block for hyperbolic decision trees.

\subsubsection{Combining classes}

The default \texttt{HoroSplitter} uses a one-versus-rest approach to deal with multi-class data. As a result, it can only find one-versus-rest splits. Nonetheless, a horosphere split with a high information gain could have more than one class in either partition. To find good splits with multiple classes while avoiding searching over all possible combinations, only the most promising combinations should be evaluated. For this, we design a heuristic that transforms the multi-class problem into a hierarchical set of binary problems by iteratively grouping classes into two hyperclasses based on their lowest common ancestor (LCA). Then, these binary combinations are added to the pool of one-versus-rest experiments and evaluated for their information gain. 

\begin{figure}[t]
  \centering
  \setlength\tabcolsep{15pt}
  \begin{tabular}{cc}
    \includegraphics[width=0.275\linewidth]{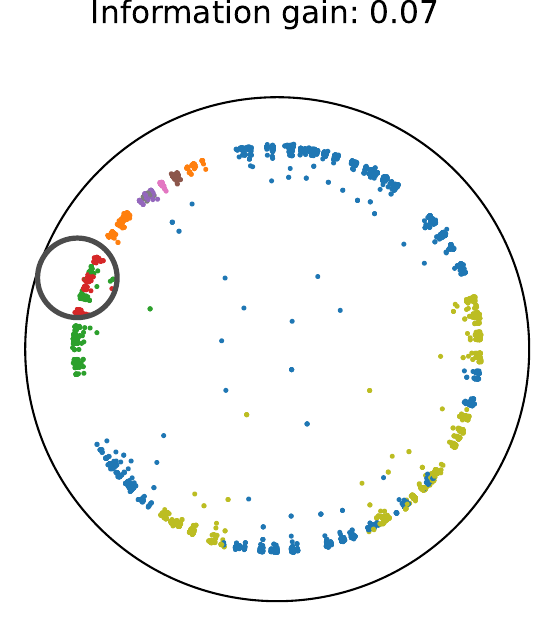} &
    \includegraphics[width=0.275\linewidth]{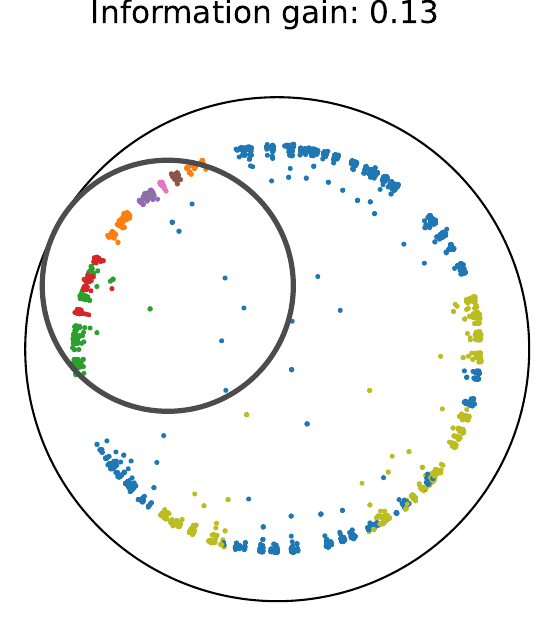} \\
    (a) & (b) \\
  \end{tabular}
    \caption{\textbf{Information gain from hierarchical splits.} The best (a) one-versus-rest split and (b) split found with our hyperclass heuristic. We find splits with a higher information gain by considering splits where more than one class is seen as positive.} % : 0.07 versus 0.13}
    \label{fig:hyper}
\end{figure}

Recall the connection of hyperbolic space to graph-theoretical trees. In a tree, similar leaf nodes lie in a small subtree and have their LCA at a high depth. In contrast, dissimilar nodes will have their LCA close to the root of the tree. We exploit the analog of this property in hyperbolic space to cluster together the most similar classes first, as similar classes are likely to be capturable by a single horosphere.

We represent each cluster by its hyperbolic mean. As computing the average in the Poincar\'e model involves the computationally expensive Fr\'echet mean \citep{lou2020differentiating}, we use the Einstein midpoint after transforming the data to Klein coordinates instead. Specifically, data is transformed into Klein space $\mathds{K}$ with 
\begin{equation}
\label{eq:poincaretoklein}
    \x_\mathds{K} = \frac{2\x_\mathds{B}}{1 + \|\x_\mathds{B}\|^2},
\end{equation}
and we compute the average $\cl{y}$ for class $y$ using
\begin{equation}
\label{eq:hypaverage}
    \cl{y} = \frac{\sum_{i=1}^N \mathds{1}_{[y_i=y]}\gamma_i\x_i}{\sum_{i=1}^N \gamma_i},
\end{equation}
where $\gamma_i = \frac{1}{\sqrt{1 - \|\x_i\|^2}}$. Afterward, the mean is transformed back to the Poincar\'e ball with 
\begin{equation}
\label{eq:kleintopoincare}
    \x_\mathds{B} = \frac{\x_\mathds{K}}{1 + \sqrt{1 - \|\x_\mathds{K}\|^2}}.
\end{equation}

We compute all pairwise similarities and iteratively group the two most similar classes together, where the similarity is based on the LCA of the means of the two classes. The LCA for two points in hyperbolic space is the point that is closest to the origin on the geodesic between them~\citep{chami2020trees}. As a result, the distance of the LCA from the origin can be seen as a similarity measure. Given two means $\cl{i}$ and $\cl{j}$, we can calculate their similarity by
\begin{equation}
\label{eq:similarity}
    \textrm{sim}(\cl{i},\cl{j}) = 2\tanh^{-1}\left(\sqrt{R^2 + 1} - R\right)
\end{equation}
with \begin{equation}
    R = \sqrt{\left(\frac{\|\cl{i}\|^2 + 1}{2\|\cl{i}\|\cos(\alpha)}\right)^2-1}
\end{equation}
and
\begin{equation}
    \alpha = \tan^{-1}\left(\frac{1}{\sin(\theta)}\left(\frac{\|\cl{i}\|(\|\cl{j}\|^2 + 1)}{\|\cl{j}\|(\|\cl{i}\|^2 + 1)} - \cos(\theta)\right)\right),
\end{equation}
where $\theta$ is the angle between $\cl{i}$ and $\cl{j}$. We repeat this until only two hyperclasses are left, for $K-2$ iterations total. As a result, we only evaluate $K + K - 2$ horospheres per node. We show an example where evaluating hyperclasses is beneficial in Fig.~\ref{fig:hyper}. 

\begin{figure}[t]
  \centering
    \includegraphics[width=0.6\linewidth]{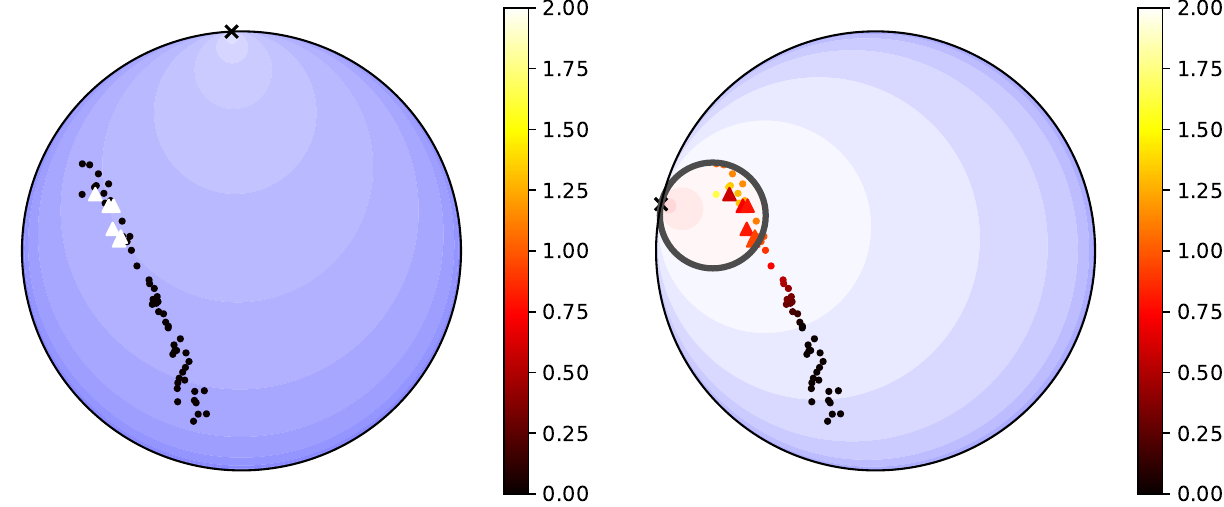} \\
    \caption{\textbf{Importance of class balancing.} The optimal solution on a binary classification problem found (left) without and (right) with class-balancing. The left horosphere has a large negative $\mathbf{b}$ and does not split the data. In contrast, the right horosphere has a higher loss but a positive information gain. Class labels are given by marker shape. The points and color bar are colored by loss; the background is colored by distance to the horosphere. }
    \label{fig:imbalanced}
\end{figure}

\subsubsection{Imbalanced data}
SVM-based methods are known to behave poorly on imbalanced data, and \texttt{HoroSplitter} is no exception. We address this by proposing a class-balanced version of the \texttt{HoroSplitter}.
Specifically, the problem comes from \eqref{eq:svmloss} finding an optimal solution that does not split the data and has an information gain of zero as a result. As an example, consider the binary problem in Figure~\ref{fig:imbalanced}. The optimal horosphere has its ideal point on the top of the Poincar\'e ball, a $\mu$ very close to 0 and $o$ very close to 1; a result of the loss, in the case of a large imbalance, almost reducing to the degenerate form
\begin{equation}
    \boldsymbol\ell_{\textrm{imbalanced}}(\mu,\w,o;D) \approxeq C\sum^{N}_{i=1}\max(0, 1 + y_i).
\end{equation}

%\emph{i.e.}, approximately 0 for all samples of the negative class and approximately 2 for all samples of the positive class. Thus, in cases where the negative class greatly outnumbers the positive class, the optimal solution finds a very small horosphere for which samples from the positive class get a slightly less confident prediction to be the negative class than samples from the negative class.

We remedy this limitation by incorporating class-balancing into \texttt{HoroSplitter}'s loss function~\citep{cui2019class}. We rewrite the loss function in \eqref{eq:svmloss} as 
\begin{equation}
\label{eq:cbsvm}
    \boldsymbol\ell(\mu,\w,o;D) = \frac{1}{2}\mu^2 + C\sum^{N}_{i=1}\frac{1 - \beta}{1 - \beta^{n_y}}\max(0, 1 - y_i(\mu B_\w^{-1}(\x_i) - o)),
\end{equation}
where $n_y$ gives the number of training samples belonging to class $y$ and $\beta$ is a hyperparameter. By adding more emphasis on the loss for samples of the minority class, we can find horospheres that split the data even in imbalanced cases. 

\paragraph{Training and inference}

We build horospherical decision trees (HoroDT) by repeatedly applying the \texttt{HoroSplitter}. This process is repeated until all nodes are pure, \emph{i.e.} contain one class, or a stopping criterion is met. Example stopping criteria include stopping when a node reaches a certain number of samples or no split with an information gain higher than a threshold can be found.
By combining multiple HoroDTs into an ensemble, we form a horospherical Random Forest (HoroRF). Each tree is trained on a randomly sampled (with replacement) subset of the data, and a subsample of the features is considered at every split. Predictions are made by a majority vote among the HoroDTs.

\section{Experiments}

\subsection{Datasets}

We evaluate HoroRF on two canonical hyperbolic classification benchmarks: WordNet subtree and network node classification~\citep{cho2019large,fan2023horocycle}. Additionally, we introduce three new multi-class WordNet experiments, which we will detail first.

\paragraph{Multi-class WordNet}
We propose three new multi-class WordNet experiments aimed at evaluating three distinct qualities. In the first experiment, the task is to classify samples into one of multiple subtrees with the same parent. In the second experiment, we pick nested subtrees, and samples must be classified into the smallest subtree they belong to. The third experiment combines the previous experiments with nested classes of multiple subtrees. Unlike previous benchmarks, these are difficult multi-class experiments on large, imbalanced datasets. We provide the full details of the experiments in the appendix. 
The datasets are split into train, validation, and test sets with a ratio of 60:20:20. A hyperparameter search is done on the validation set, and the macro-f1 score over three trials on the test set is reported. 

\paragraph{WordNet} We run binary classification experiments on the nouns of WordNet~\citep{fellbaum2010princeton}, embedded into hyperbolic space using hyperbolic entailment cones~\citep{ganea2018hyperbolic}. The goal is to identify whether a sample belongs to a semantic category (\emph{i.e.},~a subtree of the hyperbolic embeddings). Following~\citep{fan2023horocycle}, we split the data into 80\% for training and 20\% for testing and report the best results from a grid search over three runs in AUPR. We find that three of the larger subtrees typically used for these experiments (\texttt{animal.n.01}, \texttt{group.n.01}, \texttt{mammal.n.01}) can be solved near-perfectly with random forests, both Euclidean and hyperbolic. Therefore, we add experiments on two additional subtrees, \texttt{occupation.n.01} and \texttt{rodent.n.01}, which are smaller and more difficult subtrees for classification purposes.

\paragraph{Networks} We run node classification experiments on four network datasets embedded into hyperbolic space. These networks are (1) \karate~\citep{zachary1977information}, a network of 34 members of a karate club split into two factions, (2) \polblogs~\citep{adamic2005political}, a network of 1224 hyperlinks in political blogs split into two communities, (3) football~\citep{girvan2002community}, a network of 115 colleges linked by football games split into 12 leagues, and (4) \polbooks\footnote{http://www-personal.umich.edu/~mejn/netdata/}, a network of 105 books split into three affiliations.

% We run node classification experiments on four real-world network datasets embedded into hyperbolic space: \karate~(34 nodes, 2 classes; \citet{zachary1977information}), \polblogs~(1224 nodes, 2 classes; \citet{adamic2005political}), football~(115 nodes, 12 classes; \citet{girvan2002community}), and \polbooks\footnote{http://www-personal.umich.edu/~mejn/netdata/}~(105 nodes, 3 classes).

We use the public embeddings provided by \citet{cho2019large}. Each dataset has five sets of embeddings obtained using the method of \citet{chamberlain2017neural}. Following previous work~\citep{fan2023horocycle}, we run a grid search with 5-fold stratified cross-validation and report the best average micro-f1 score over five seeds, using a different embedding for each seed. 

\subsection{Evaluation}

\paragraph{Baselines}
We evaluate HoroRF against six baselines, split into two categories. 
The first set covers hyperbolic classifier baselines, namely hyperbolic multiple logistic regression (HypMLR; \citet{ganea2018nns}) and the state-of-the-art hyperbolic SVM, HoroSVM~\citep{fan2023horocycle}.
The second set of baselines comprises Euclidean counterparts of the hyperbolic classifiers, namely a linear SVM (LinSVM), an SVM with an RBF kernel (RBFSVM; \citep{cortes1995support}), an axis-aligned random forest (RF)~\citep{breiman2001random}, and an oblique random forest (OblRF) with a similar set-up to HoroRF. The OblRF uses a cost-balanced linear SVM to find optimal splits (see \emph{e.g.}~\citep{do2010classifying}). If the SVM fails to converge, it defaults to using the best axis-aligned split. It finds hyperclasses by iteratively merging the two classes with the closest means. 

As is common practice~\citep{cho2019large,fan2023horocycle}, the Euclidean baselines are run on the same embeddings as the hyperbolic methods, which we also find to perform better compared to mapping the embeddings back to Euclidean space first. Moreover, as the SVMs, HypMLR, and HoroSVM baselines are negatively affected by the class imbalance in the WordNet experiments, we report cost-balanced results on those benchmarks, where the loss is weighted inversely proportional to the class frequency.

\paragraph{Implementation Details}
We use 100 trees for all tree-based methods. We use the Gini impurity to calculate the information gain. At every node, we set~$C$ to $2^n$ where $n$ is randomly sampled from $\{-3, -2, \ldots, 5\}$. We stop splitting when a node reaches a certain number of samples~$m$. In the case of ties in information gain between possible splits, a random split is selected. If the \texttt{HoroSplitter} fails to converge, we choose the best horosphere from a small number of random ideal points. We set~$m$ and the class-balancing hyperparameter~$\beta$ via grid search. Further details are given in the appendix.

\subsection{Results}

\paragraph{WordNet} The results for the WordNet subtree classification experiments are shown in Tab.~\ref{tab:wordnet}. The Euclidean SVMs are unable to handle the large imbalance in the datasets despite the loss re-weighting. The random forest approaches are better equipped to deal with this type of data. The results show that HoroRF outperforms the Euclidean tree-based methods in most cases. The differences are most pronounced in the hardest experiments, \occupation~and \rodent, while we find the tree-based methods to perform similarly on the saturated experiments with close to perfect scores, i.e., \animal, \group, and \mammal. Overall, HoroRF performs better than the well-known Euclidean SVMs and random forest approaches.

Compared to the hyperbolic classifiers, we find that both the hyperbolic MLR and SVM baselines struggle to deal with the imbalanced data, despite the loss re-weighting. Neither are able to match the performance of HoroRF. We argue this is the case as real-world data cannot be distinguished with a single horosphere, but requires recursive horospherical separation as done in HoroRF. We conclude from the WordNet experiment that HorRF is a competitive classifier even in the binary setting.

\begin{table*}[t]
\centering
\resizebox{\linewidth}{!}{
% {\color{blue}
\begin{tabular}{lcccccccc}
\toprule
& \multirow{2}{*}{Animal} & \multirow{2}{*}{Group} & \multirow{2}{*}{Worker} & \multirow{2}{*}{Mammal} & \multirow{2}{*}{Tree} & \multirow{2}{*}{Solid} & \multirow{2}{*}{Occupation}  & \multirow{2}{*}{Rodent}\\
 &&&&&&&\\
\midrule
\rowcolor{mygray} \textbf{Euclidean} & & & & & & & & \\
LinSVM & 2.5\small{$\pm0.0$} & 6.0\small{$\pm0.6$} & 0.7\small{$\pm0.0$} & 0.7\small{$\pm0.0$} & 0.6\small{$\pm0.0$} & 0.8\small{$\pm0.0$} & 0.2\small{$\pm0.0$} & 0.1\small{$\pm0.0$} \\
RBFSVM & 2.5\small{$\pm0.0$} & 6.0\small{$\pm0.6$} & 0.7\small{$\pm0.0$} & 0.7\small{$\pm0.0$} & 0.6\small{$\pm0.0$} & 0.8\small{$\pm0.0$} & 0.2\small{$\pm0.0$} & 0.1\small{$\pm0.0$} \\
RF & \textbf{98.6}\small{$\pm$0.2} &  96.6\small{$\pm$0.4} & 70.0\small{$\pm3.1$} & \textbf{99.3}\small{$\pm$0.3} & 75.8\small{$\pm2.0$} & 92.5\small{$\pm$ 1.2} & 59.8\small{$\pm1.1$} & 34.5\small{$\pm3.0$}\\
OblRF & 98.4\small{$\pm$0.1} & \textbf{96.7} \small{$\pm$0.3} & 73.3\small{$\pm$1.3} & 98.8\small{$\pm$0.3} & 76.2\small{$\pm$3.2} & 92.9\small{$\pm$1.7} & 65.0\small{$\pm$2.4} & 38.2\small{$\pm$0.9}\\
\midrule
\rowcolor{mygray} \textbf{Hyperbolic} & & & & & & & & \\
HypMLR & 3.2 \small{$\pm$0.1} & 6.5 \small{$\pm$0.0} & 1.2 \small{$\pm$0.2} & 0.9 \small{$\pm$0.0} & 1.3\small{$\pm$0.8} & 1.0 \small{$\pm$0.2} & 0.9\small{$\pm$0.9} & 0.4 \small{$\pm$0.4}\\
HoroSVM & 92.4\small{$\pm1.3$} & 71.0\small{$\pm0.5$} & 39.9\small{$\pm3.6$} & 89.8\small{$\pm0.9$} & 41.5\small{$\pm1.0$} & 65.7\small{$\pm0.9$} & 13.0\small{$\pm1.5$} & 13.6\small{$\pm2.0$} \\
% HypRF & 98.5\small{$\pm$0.2} & 96.6\small{$\pm$0.1} & 71.5\small{$\pm$0.9} & 99.1\small{$\pm$0.0} & 77.5\small{$\pm$2.1} & 92.2\small{$\pm$0.5} & 53.9\small{$\pm$4.4} & 40.3\small{$\pm$1.6}\\
\emph{HoroRF} & \underline{98.5}\small{$\pm$0.2} & \underline{96.6}\small{$\pm$0.4} & \underline{\textbf{73.4}}\small{$\pm$ 1.7} & \underline{\textbf{99.3}}\small{$\pm$0.4} &  \underline{\textbf{76.6}}\small{$\pm$ 3.8} & \underline{\textbf{93.5}}\small{$\pm$ 1.3} & \underline{\textbf{65.6}}\small{$\pm$ 1.5} & \underline{\textbf{39.0}}\small{$\pm$ 2.4} \\
\bottomrule
\end{tabular}
% }
}
\caption{\textbf{Comparative evaluation on binary WordNet experiments.} We follow the experimental protocol of~\cite{fan2023horocycle} and report the mean and standard deviation of the AUPR over three runs. \underline{Underlined} gives the best hyperbolic method, \textbf{Bold} denotes the best method overall. On aggregate across the experiments, HoroRF obtains the best performance.}
\label{tab:wordnet}
\end{table*}

\paragraph{Multi-class WordNet}

The results for the multi-class WordNet experiments are shown in Table~\ref{tab:wordnet_multi}, which paint a similar picture as the previous experiments. For the Euclidean methods, the tree-based methods outperform the SVMs, with OblRF, on average, having a slight edge over the axis-aligned random forest. HoroRF outperforms the SVMs and axis-aligned RF in all three cases, and despite reaching an equal score in the first experiment, it outperforms OblRF in the remaining two. Compared to the other hyperbolic classifiers, HoroRF reaches far higher performance on these complex imbalanced multi-class experiments, on which both HypMLR and HoroSVM struggle greatly. 

\begin{table}[t!]
\centering
% \resizebox{\linewidth}{!}{
% {\color{blue}
\begin{tabular}{lccc}
\toprule
& Same & \multirow{2}{*}{Nested} & \multirow{2}{*}{Both} \\
& level\\
\midrule
\rowcolor{mygray} \textbf{Euclidean} & & &\\
LinSVM & 48.8 \small{$\pm$0.9} & 59.7 \small{$\pm$0.5} & 35.6 \small{$\pm$0.2} \\
RBFSVM & 80.0 \small{$\pm$1.8} & 89.8 \small{$\pm$1.1} & 70.8 \small{$\pm$0.6} \\
RF & 89.7\small{$\pm$0.7} & 91.7\small{$\pm$0.3} & 81.5\small{$\pm$1.5} \\
OblRF &  \textbf{91.3}\small{$\pm$0.6} & 93.0\small{$\pm$0.2} & 81.3\small{$\pm$1.4}\\
\midrule
\rowcolor{mygray} \textbf{Hyperbolic} & & &\\
HypMLR & 29.0 \small{$\pm$4.5} & 57.6 \small{$\pm$0.5} & 35.9 \small{$\pm$6.7} \\
HoroSVM & 50.2 \small{$\pm$2.2} & 56.0 \small{$\pm$0.7} & 35.2 \small{$\pm$0.5} \\
% HypRF & 93.8 \small{$\pm$0.8} & 93.8 \small{$\pm$0.8} & 84.7 \small{$\pm$2.1}\\
\emph{HoroRF} & \underline{\textbf{91.3}}\small{$\pm$0.3} & \underline{\textbf{93.3}}\small{$\pm$1.1} & \underline{\textbf{81.9}}\small{$\pm$1.5} \\
\bottomrule
\end{tabular}
% }
% }
\caption{\textbf{Comparative evaluation on multi-class WordNet.} We run a grid search to determine the optimal configuration on the validation set and report results over three trials in macro-f1 score on the test set. \underline{Underlined} gives the best hyperbolic method, \textbf{Bold} denotes the best method overall. HoroRF performs best in all settings.}
\label{tab:wordnet_multi}
\end{table}

\paragraph{Networks} The results on the network datasets are shown in Table~\ref{tab:networks}. Our HoroRF outperforms the Euclidean methods in all cases, besides matching the SVMs on \karate, showing that the choice of horospheres over hyperplanes results in better performance on data lying in hyperbolic space. 

As for the hyperbolic methods, HypMLR is outclassed by both HoroRF and HoroSVM in all cases except \football. HoroSVM and HoroRF perform similarly in the binary experiments, with a slight edge for HoroRF on \polblogs. In contrast, for the multi-class experiments \football~and \polbooks, HoroRF outclasses HoroSVM by 4.0 and 0.6 macro f-1, respectively, confirming the benefit of HoroRF over HoroSVM in more complex cases. Overall, we find that HoroRF is the most consistent and best-performing classifier across all benchmarks.

\begin{table}[t!]
% \resizebox{\linewidth}{!}{
\centering
% {\color{blue}
\begin{tabular}{lcccc}
\toprule
& \multicolumn{2}{c}{Binary} & \multicolumn{2}{c}{Multi-class}\\
\cmidrule(lr){2-3} \cmidrule(lr){4-5}
& Karate & Polbooks & Football & Polblogs\\
\midrule
\rowcolor{mygray} \textbf{Euclidean} & & & & \\
LinSVM & \textbf{95.4}\small{$\pm$2.3} & 92.4\small{$\pm$0.3} & 33.2\small{$\pm$5.1} & 85.5\small{$\pm$0.9}\\
RBFSVM & \textbf{95.4}\small{$\pm$2.3} & 92.4\small{$\pm$0.3} & 35.5\small{$\pm$4.7} & 84.4\small{$\pm$1.9}\\
RF & 94.3\small{$\pm$3.1} & 92.1\small{$\pm$0.3} & 36.2\small{$\pm$4.9} & 85.1\small{$\pm$2.1}\\
OblRF & 94.8 \small{$\pm$2.2} & 92.1 \small{$\pm$0.3} & 36.7 \small{$\pm$2.7} & 84.4 \small{$\pm$1.3}\\
\midrule
\rowcolor{mygray} \textbf{Hyperbolic} & & & & \\
HypMLR & 93.1\small{$\pm3.4$} & 90.9\small{$\pm0.7$} & \underline{\textbf{40.2}}\small{$\pm2.5$}  & 81.5\small{$\pm1.5$}\\
HoroSVM & \textbf{\underline{95.4}}\small{$\pm$2.3} & 92.4\small{$\pm$0.2} & 34.3\small{$\pm$1.8} & 85.3\small{$\pm$0.8} \\
% HypRF & 93.0 \small{$\pm$4.4} & 91.7 \small{$\pm$0.4} & 33.0 \small{$\pm$2.3} & 84.4 \small{$\pm$1.8} \\
\emph{HoroRF} & \underline{\textbf{95.4}}\small{$\pm$2.3} & \underline{\textbf{92.5}}\small{$\pm$0.3} & 38.3\small{$\pm$1.8} & \underline{\textbf{86.1}}\small{$\pm$1.0}\\
\bottomrule
\end{tabular}
% }
% }
\caption{\textbf{Comparative evaluation on network datasets.} We follow the experimental protocol of~\cite{fan2023horocycle} and report the mean and standard deviation of the micro-f1 score over five trials of 5-fold stratified cross-validation. \underline{Underlined} gives the best hyperbolic method, \textbf{Bold} denotes the best method overall. HoroRF obtains the best performance on aggregate over the datasets.}
\label{tab:networks}
\end{table}

\subsection{Ablations and Visualizations}

\paragraph{Hierarchical Classification} Hyperbolic methods have been proven successful when evaluated on a wide range of hierarchical tasks~\cite{cao2020hypercore,ghadimi2021hyperbolic,suris2021learning,tifrea2018poincar}. Here, we aim to verify our hypothesis that random forests are well-suited for hyperbolic space by comparing HoroRF with other SOTA hyperbolic classifiers on hierarchical classification, with metrics explicitly designed to evaluate their success in capturing the hierarchy.

We evaluate the methods using CIFAR10~\cite{krizhevsky2009learning} and STL10~\cite{coates2011analysis} using the CIFAR10 hierarchy from~\cite{sebastian2023leveraging}. Note that STL10 has the same classes as CIFAR10, with the only difference being the substitution of \texttt{frog} for \texttt{monkey}. As such, we maintain the hierarchy from~\cite{sebastian2023leveraging} for STL10 but modify it by removing \texttt{frog} and incorporating \texttt{monkey} into the mammal subtree. 

We embed the images into 9-dimensional space by training a hyperbolic prototype ResNet-18 with prototypes uniformly distributed over the hyperbole~\cite{kasarla2022maximum} and train the classifiers on these 9-dimensional image features. We obtain hyperbolic embeddings for 3600 training, 900 validation, and 1000 test samples. To compare the classifiers, we run a grid search on the training set and select the optimal configuration based on validation performance. Then, we report the mistake severity and hierarchical distance@1 on the test set~\cite{garg2022learning}. These metrics both use the LCA distance, computed by dividing the height of the LCA of the predicted and ground-truth class by the tree height. The mistake severity is the average of this distance for all misclassified samples, whereas the HD@1 computes the average distance for all the samples~\cite{karthik2021no}.

From the results in Tab.~\ref{tab:vision}, HoroRF is the best hyperbolic method, confirming its ability to model the hierarchy effectively and the benefits of our generalization of random forests to hyperbolic space.

\begin{table}[t!]
% \resizebox{\linewidth}{!}{
\centering
\begin{tabular}{lcccc}
\toprule
& \multicolumn{2}{c}{CIFAR10} & \multicolumn{2}{c}{STL10}\\
& Mis. Sev. ($\downarrow$) & HD@1 ($\downarrow$) & Mis. Sev. ($\downarrow$) & HD@1 ($\downarrow$) \\
\midrule
% \rowcolor{mygray} \textbf{Euclidean} & & & & & & \\
% LinSVM & \textbf{89.8}\small{$\pm$1.7}& 59.6\small{$\pm$2.0} & 9.4\small{$\pm$1.3} & 57.5\small{$\pm$2.0} & 53.7\small{$\pm$1.2} & 22.8\small{$\pm$1.0}\\
% RBFSVM & 89.7\small{$\pm$1.9} & 57.6\small{$\pm$2.9} & \textbf{9.2}\small{$\pm$1.5} & \textbf{58.5}\small{$\pm$1.9} & 53.7\small{$\pm$1.4} &  \textbf{22.2}\small{$\pm$0.9} \\
% RF & 88.3\small{$\pm$2.3} & 57.6\small{$\pm$0.8} & 9.9\small{$\pm$1.4} & 56.4\small{$\pm$1.9} & 54.0\small{$\pm$1.0} & 22.7\small{$\pm$0.7}\\
% OblRF & 89.1\small{$\pm$1.8} & 58.8\small{$\pm$3.1} & 9.7\small{$\pm$1.4}  & 58.4\small{$\pm$1.3} & 54.1\small{$\pm$0.6} & 22.5\small{$\pm$0.7}\\
% \midrule
% \rowcolor{mygray} \textbf{Hyperbolic} & & & & & & \\
HypMLR &  60.8\small{$\pm$1.7} & 10.2\small{$\pm$0.0} &  54.2\small{$\pm$2.8} & 22.4\small{$\pm$0.3} \\
HoroSVM & 59.9\small{$\pm$0.1} & 12.5\small{$\pm$0.0} & 54.4\small{$\pm$1.0} & 26.0\small{$\pm$0.3} \\ 
\emph{HoroRF} & \textbf{58.2}\small{$\pm$0.3} & \textbf{9.9}\small{$\pm$0.2} & \textbf{53.4}\small{$\pm$0.2} & \textbf{22.3}\small{$\pm$0.1}\\
\bottomrule
\end{tabular}
% }
\caption{\textbf{Evaluating hyperbolic classifiers for hierarchical classification.} We run a grid search to determine the optimal configuration on the validation set and report the mean and standard deviation of the mistake severity and hierarchical distance@1 on the test set over three seeds. \textbf{Bold} denotes the best method. HoroRF performs best.} 
\label{tab:vision}
\end{table}

\paragraph{Ablating hyperclasses and balancing.}

\begin{table}
\begin{center}
% \resizebox{\linewidth}{!}{
\begin{tabular}{cccc}
\toprule
 Ideal & Hyper- & Class- & \multirow{2}{*}{f1} \\
 Points & classes & balanced & \\
\midrule
 Axis-aligned & \xmark & \xmark & 9.9\\
 \texttt{HoroSplitter} & \xmark & \xmark & 35.7\\
 \texttt{HoroSplitter} & \xmark & \cmark & 36.5\\
 \texttt{HoroSplitter} & \cmark & \xmark & 37.4\\
 \texttt{HoroSplitter} & \cmark & \cmark & \textbf{38.3} \\
 \bottomrule
\end{tabular}
% }
\end{center}
\caption{\textbf{Ablating the \texttt{HoroSplitter}} in HoroRF on the \football~dataset. We follow the experimental protocol of~\cite{fan2023horocycle} and report the mean and standard deviation of the micro-f1 score over five trials of 5-fold stratified cross-validation. Our splitting function, hyperclass-based aggregation, and class balancing are all important for effective hyperbolic random forests.}
\label{tab:ablation}
\end{table}

We perform an ablation study to validate our choices on the \football~dataset. From Tab.~\ref{tab:ablation}, we find a large benefit in using our \texttt{HoroSplitter} to find splits instead of enumerating all possible horospheres at axis-aligned ideal points. Furthermore, we show how both our hyperclass and class-balancing additions improve upon the base \texttt{HoroSplitter}, and their combination enables us to reach the best performance.

\paragraph{Visualization \& synthetic evaluation}

We build a synthetic dataset following \citet{ganea2018hyperbolic} by creating a synthetic tree of depth six with a branching factor of four and embedding it into hyperbolic space with hyperbolic entailment cones. We then partition the nodes into five classes from two depths. Using this dataset, we visualize the splits of HoroRF, HoroRF without hyperclasses or class-balancing, OblRF on the hyperbolic embeddings, and OblRF on the Euclidean embeddings in Fig.~\ref{fig:vis}. We additionally verify which method is the most successful at various depths in Tab.~\ref{tab:depth}, where we find that HoroRF consistently reaches high performance and is especially effective with a single split.

\begin{figure*}[t]
  \centering
  \begin{tabular}{c}
   \includegraphics[width=0.9\linewidth]{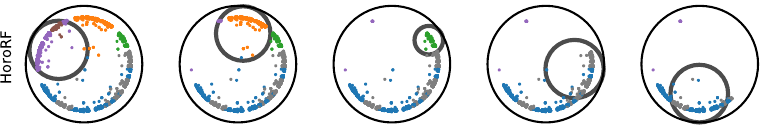} \\
   \includegraphics[width=0.9\linewidth]{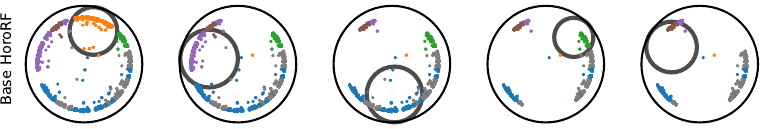} \\
   \includegraphics[width=0.9\linewidth]{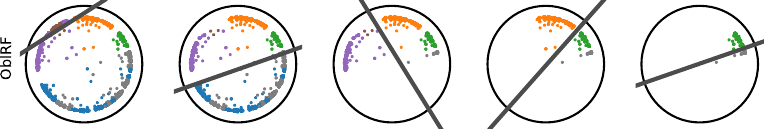} \\
   \includegraphics[width=0.9\linewidth]{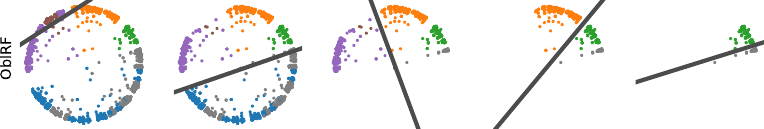} \\
  \end{tabular}
    \caption{\textbf{Visualizing HoroRF \& OblRF splits.} We show five splits for HoroRF, a basic version of HoroRF without class-balancing and hyperclasses, OblRF on the original hyperbolic embeddings, and OblRF after mapping the data to Euclidean space. The split with the highest impurity is chosen at every level to visualize the next level. Splitting in hyperbolic space with balanced and multi-level horospheres obtains good splits for hyperbolic classification.}
    \label{fig:vis}
\end{figure*}

\begin{table}[t]
\centering
% \resizebox{\linewidth}{!}{
\begin{tabular}{lccccc}
\toprule
Depth & \multirow{2}{*}{1} & \multirow{2}{*}{2} & \multirow{2}{*}{3} & \multirow{2}{*}{10}  & \multirow{2}{*}{No limit}\\
& \\
\midrule
OblRF Euc & 34.5\small{$\pm$0.1} & \textbf{68.8}\small{$\pm$2.5} & 83.3\small{$\pm$0.0} & \textbf{99.3}\small{$\pm$0.0} & \underline{98.9}\small{$\pm$0.0}\\
OblRF Hyp & 33.6\small{$\pm$0.7} & 65.6\small{$\pm$1.3} & 83.2\small{$\pm$0.3} & \underline{99.1}\small{$\pm$0.0} & \underline{98.9}\small{$\pm$0.0}\\
Base HoroRF & \underline{60.8}\small{$\pm$0.0} & \underline{67.8}\small{$\pm$4.9} & \underline{83.6}\small{$\pm$1.6} & \underline{99.1}\small{$\pm$0.0} & \textbf{99.1}\small{$\pm$0.3}\\
HoroRF & \textbf{63.1}\small{$\pm$3.2} & 67.4\small{$\pm$5.1} & \textbf{85.5}\small{$\pm$1.8} & \textbf{99.3}\small{$\pm$0.0} & \underline{98.9}\small{$\pm$0.3}\\
\bottomrule
\end{tabular}
% }
\caption{\textbf{Effect of tree depth on performance on the synthetic dataset.} We run a grid search to determine the optimal configuration on the validation set and report results over three trials in micro-f1 score on the test set. \textbf{Bold} denotes the best method, \underline{underlined} the second best. Splitting in hyperbolic space with balanced and multi-level horospheres consistently reaches high performance.}
\label{tab:depth}
\end{table}

% \begin{table}[t]
% \centering
% % \resizebox{\linewidth}{!}{
% {\color{blue}\begin{tabular}{lccccc}
% \toprule
% & \multirow{2}{*}{1} & \multirow{2}{*}{2} & \multirow{2}{*}{5}  & \multirow{2}{*}{10} & \multirow{2}{*}{Max.}\\
% & \\
% \midrule
% OblRF Euc & 41.1\small{$\pm$7.9} & 61.4\small{$\pm$1.0} & 86.7\small{$\pm$0.6} & 89.3\small{$\pm$2.4} & 90.9\small{$\pm$1.7}\\
% OblRF Hyp & 33.9\small{$\pm$0.7} & 67.0\small{$\pm$1.2} & \textbf{92.1}\small{$\pm$0.7} & 88.6\small{$\pm$2.7} & 87.7\small{$\pm$0.7}\\
% Base HoroRF & \textbf{63.8}\small{$\pm$4.4} & \underline{71.9}\small{$\pm$0.3} & 87.5\small{$\pm$2.9} & \underline{98.8}\small{$\pm$0.1} & \textbf{99.1}\small{$\pm$3.0}\\
% HoroRF & \underline{61.3}\small{$\pm$6.8} & \textbf{73.6}\small{$\pm$1.8} & \underline{89.1}\small{$\pm$3.4} & \textbf{99.0}\small{$\pm$0.2} & \underline{98.9}\small{$\pm$3.0}\\
% \bottomrule
% \end{tabular}}
% % }
% \caption{\ld{\textbf{Effect of tree depth on performance on the synthetic dataset.} We run a grid search to determine the optimal configuration on the validation set and report results over three trials in micro-f1 score on the test set. \textbf{Bold} denotes the best method, \underline{underlined} the second best. Splitting in hyperbolic space with balanced and multi-level horospheres consistently reaches high performance.}}
% \label{tab:depth}
% \end{table}

\paragraph{Reducing runtime}

\begin{figure*}[t]
  \centering
  \setlength\tabcolsep{15pt}
  \begin{tabular}{cc}
   \includegraphics[width=0.3\linewidth]{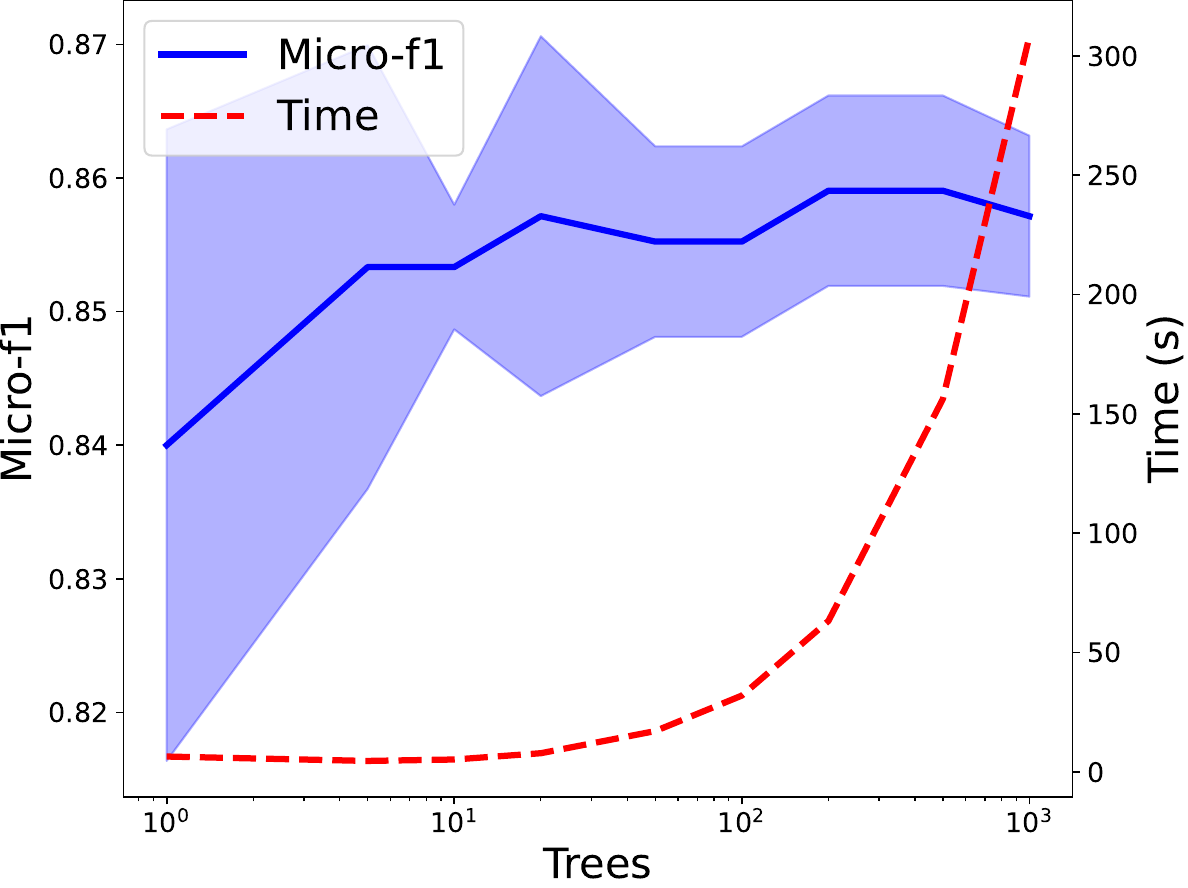} &
   \includegraphics[width=0.3\linewidth]{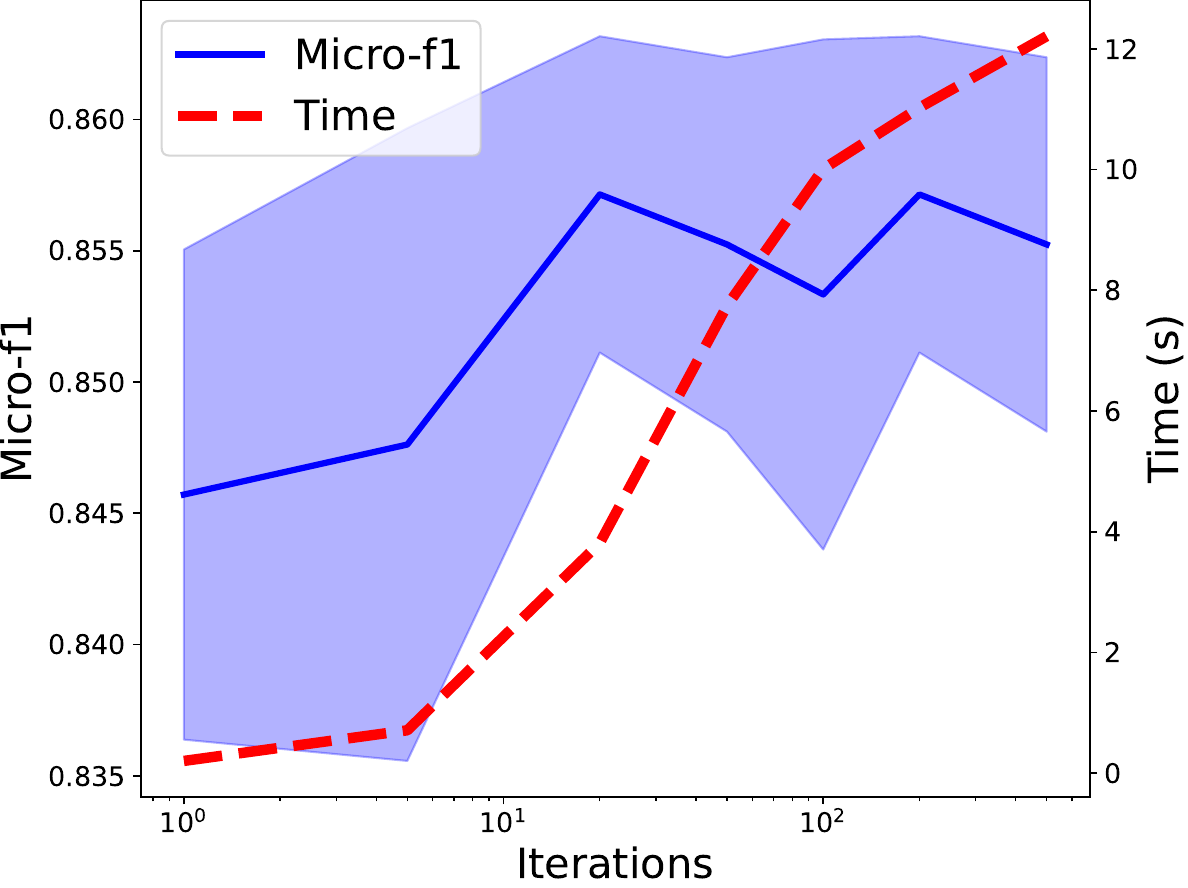} \\
  \end{tabular}
    \caption{\textbf{Analyzing ways to reduce the computational time of HoroRF on \polbooks.} In (a), we show that the optimal performance is already reached with around 20 trees. In (b), we show that by reducing the number of optimization iterations of HoroSVM, we can further reduce computational time while maintaining high performance.}
    \label{fig:compl}
\end{figure*}

A limitation of HoroRF is its runtime on large datasets. While the \texttt{HoroSplitter} scales linearly with the number of samples as HoroSVM has a linear time complexity, it is applied multiple times per node and repeated for every node. We designed and conducted two experiments showing how to mitigate the computational time needed to obtain good results: reducing the number of trees and lowering the number of HoroSVM optimization iterations. From the results in Fig.~\ref{fig:compl}, we find that both the number of trees and HoroSVM optimization iterations can be greatly reduced without sacrificing performance. Lowering the number of HoroSVM iterations is a logical way to reduce computational time, as we are not interested in finding the exact optimal horosphere; a solution close to it suffices for our purpose and might even benefit HoroRF due to the increased variation in splits.

\section{Conclusion}
We present HoroRF, a random forest algorithm in hyperbolic space. HoroRF constructs its trees by repeated application of the \texttt{HoroSplitter}, which finds horosphere-based splits with high information gain. We show the benefits of additional components aimed at improving performance on imbalanced and multi-class datasets. Extensive experiments on two established and one newly introduced benchmark show its effectiveness over both existing hyperbolic classifiers as well as their Euclidean counterparts.

Further advances in hyperbolic classification can easily be incorporated into the framework, as HoroRF is in no way restricted to a single type of classifier. In future work, we aim to investigate the benefits of combining multiple types of splits, for example by including hyperbolic logistic regression to provide additional split options to include next to the horosphere splits. We furthermore expect that any improvement in optimization of HoroSVM as splitting operation, akin to hyper-optimized libraries for Euclidean SVMs~\citep{chang2011libsvm}, will have direct benefits for HoroRF, especially when dealing with many classes.
%\ld{Moreover, the current choice of Horosvm for the splitting function limits the applicability of HoroRF in cases with a large number of samples or classes, due to the lack of an optimized implementation of HoroSVM, whereas hyper-optimized libraries for Euclidean SVMs are readily available~\cite{chang2011libsvm}. Reducing the computational time or HoroRF is another research direction we plan to explore.}

\bibliography{main}

\begin{thebibliography}{65}
\providecommand{\natexlab}[1]{#1}
\providecommand{\url}[1]{\texttt{#1}}
\expandafter\ifx\csname urlstyle\endcsname\relax
  \providecommand{\doi}[1]{doi: #1}\else
  \providecommand{\doi}{doi: \begingroup \urlstyle{rm}\Url}\fi

\bibitem[Adamic \& Glance(2005)Adamic and Glance]{adamic2005political}
Lada~A Adamic and Natalie Glance.
\newblock The political blogosphere and the 2004 us election: divided they blog.
\newblock In \emph{Proceedings of the 3rd international workshop on Link discovery}, pp.\  36--43, 2005.

\bibitem[Ahmad \& Lecue(2022)Ahmad and Lecue]{ahmad2022fisheyehdk}
Ola Ahmad and Freddy Lecue.
\newblock Fisheyehdk: Hyperbolic deformable kernel learning for ultra-wide field-of-view image recognition.
\newblock In \emph{Proceedings of the AAAI Conference on Artificial Intelligence}, volume~36, pp.\  5968--5975, 2022.

\bibitem[Breiman(2001)]{breiman2001random}
Leo Breiman.
\newblock Random forests.
\newblock \emph{Machine learning}, 45:\penalty0 5--32, 2001.

\bibitem[Busemann(1955)]{busemann2012geometry}
Herbert Busemann.
\newblock \emph{The geometry of geodesics}.
\newblock Academic Press, 1955.

\bibitem[Cai et~al.(2021)Cai, Zheng, Chen, Jagadish, Ooi, and Zhang]{cai2021arm}
Shaofeng Cai, Kaiping Zheng, Gang Chen, HV~Jagadish, Beng~Chin Ooi, and Meihui Zhang.
\newblock Arm-net: Adaptive relation modeling network for structured data.
\newblock In \emph{Proceedings of the 2021 International Conference on Management of Data}, pp.\  207--220, 2021.

\bibitem[Cant{\'u}-Paz \& Kamath(2003)Cant{\'u}-Paz and Kamath]{cantu2003inducing}
Erick Cant{\'u}-Paz and Chandrika Kamath.
\newblock Inducing oblique decision trees with evolutionary algorithms.
\newblock \emph{IEEE Transactions on Evolutionary Computation}, 7\penalty0 (1):\penalty0 54--68, 2003.

\bibitem[Cao et~al.(2020)Cao, Chen, Liu, Zhao, Liu, and Chong]{cao2020hypercore}
Pengfei Cao, Yubo Chen, Kang Liu, Jun Zhao, Shengping Liu, and Weifeng Chong.
\newblock Hypercore: Hyperbolic and co-graph representation for automatic icd coding.
\newblock In \emph{Proceedings of the 58th Annual Meeting of the Association for Computational Linguistics}, pp.\  3105--3114, 2020.

\bibitem[Caruana \& Niculescu-Mizil(2006)Caruana and Niculescu-Mizil]{caruana2006empirical}
Rich Caruana and Alexandru Niculescu-Mizil.
\newblock An empirical comparison of supervised learning algorithms.
\newblock In \emph{Proceedings of the 23rd international conference on Machine learning}, pp.\  161--168, 2006.

\bibitem[Chamberlain et~al.(2017)Chamberlain, Clough, and Deisenroth]{chamberlain2017neural}
Benjamin~Paul Chamberlain, James Clough, and Marc~Peter Deisenroth.
\newblock Neural embeddings of graphs in hyperbolic space.
\newblock \emph{arXiv preprint arXiv:1705.10359}, 2017.

\bibitem[Chami et~al.(2019)Chami, Ying, R{\'e}, and Leskovec]{chami2019hyperbolic}
Ines Chami, Zhitao Ying, Christopher R{\'e}, and Jure Leskovec.
\newblock Hyperbolic graph convolutional neural networks.
\newblock \emph{Advances in neural information processing systems}, 2019.

\bibitem[Chami et~al.(2020)Chami, Gu, Chatziafratis, and R{\'e}]{chami2020trees}
Ines Chami, Albert Gu, Vaggos Chatziafratis, and Christopher R{\'e}.
\newblock From trees to continuous embeddings and back: Hyperbolic hierarchical clustering.
\newblock \emph{Advances in Neural Information Processing Systems}, 33:\penalty0 15065--15076, 2020.

\bibitem[Chami et~al.(2021)Chami, Gu, Nguyen, and R{\'e}]{chami2021horopca}
Ines Chami, Albert Gu, Dat~P Nguyen, and Christopher R{\'e}.
\newblock Horopca: Hyperbolic dimensionality reduction via horospherical projections.
\newblock In \emph{International Conference on Machine Learning}, pp.\  1419--1429. PMLR, 2021.

\bibitem[Chang \& Lin(2011)Chang and Lin]{chang2011libsvm}
Chih-Chung Chang and Chih-Jen Lin.
\newblock Libsvm: A library for support vector machines.
\newblock \emph{ACM transactions on intelligent systems and technology (TIST)}, 2\penalty0 (3):\penalty0 1--27, 2011.

\bibitem[Chien et~al.(2021)Chien, Pan, Tabaghi, and Milenkovic]{chien2021highly}
Eli Chien, Chao Pan, Puoya Tabaghi, and Olgica Milenkovic.
\newblock Highly scalable and provably accurate classification in poincar{\'e} balls.
\newblock In \emph{2021 IEEE International Conference on Data Mining (ICDM)}, pp.\  61--70. IEEE, 2021.

\bibitem[Chlenski et~al.(2023)Chlenski, Turok, Moretti, and Pe'er]{chlenski2023fast}
Philippe Chlenski, Ethan Turok, Antonio Moretti, and Itsik Pe'er.
\newblock Fast hyperboloid decision tree algorithms.
\newblock \emph{arXiv preprint arXiv:2310.13841}, 2023.

\bibitem[Cho et~al.(2019)Cho, DeMeo, Peng, and Berger]{cho2019large}
Hyunghoon Cho, Benjamin DeMeo, Jian Peng, and Bonnie Berger.
\newblock Large-margin classification in hyperbolic space.
\newblock In \emph{The 22nd international conference on artificial intelligence and statistics}, pp.\  1832--1840. PMLR, 2019.

\bibitem[Coates et~al.(2011)Coates, Ng, and Lee]{coates2011analysis}
Adam Coates, Andrew Ng, and Honglak Lee.
\newblock An analysis of single-layer networks in unsupervised feature learning.
\newblock In \emph{Proceedings of the fourteenth international conference on artificial intelligence and statistics}, pp.\  215--223. JMLR Workshop and Conference Proceedings, 2011.

\bibitem[Cortes \& Vapnik(1995)Cortes and Vapnik]{cortes1995support}
Corinna Cortes and Vladimir Vapnik.
\newblock Support-vector networks.
\newblock \emph{Machine learning}, 20:\penalty0 273--297, 1995.

\bibitem[Cui et~al.(2019)Cui, Jia, Lin, Song, and Belongie]{cui2019class}
Yin Cui, Menglin Jia, Tsung-Yi Lin, Yang Song, and Serge Belongie.
\newblock Class-balanced loss based on effective number of samples.
\newblock In \emph{Proceedings of the IEEE/CVF conference on computer vision and pattern recognition}, pp.\  9268--9277, 2019.

\bibitem[Desai et~al.(2023)Desai, Nickel, Rajpurohit, Johnson, and Vedantam]{desai2023hyperbolic}
Karan Desai, Maximilian Nickel, Tanmay Rajpurohit, Justin Johnson, and Shanmukha~Ramakrishna Vedantam.
\newblock Hyperbolic image-text representations.
\newblock In \emph{International Conference on Machine Learning}, 2023.

\bibitem[Do et~al.(2010)Do, Lenca, Lallich, and Pham]{do2010classifying}
Thanh-Nghi Do, Philippe Lenca, St{\'e}phane Lallich, and Nguyen-Khang Pham.
\newblock Classifying very-high-dimensional data with random forests of oblique decision trees.
\newblock \emph{Advances in knowledge discovery and management}, pp.\  39--55, 2010.

\bibitem[Durrant \& Leontidis(2023)Durrant and Leontidis]{durrant2023hmsn}
Aiden Durrant and Georgios Leontidis.
\newblock Hmsn: Hyperbolic self-supervised learning by clustering with ideal prototypes.
\newblock \emph{arXiv preprint arXiv:2305.10926}, 2023.

\bibitem[Fan et~al.(2023)Fan, Yang, and Vemuri]{fan2023horocycle}
Xiran Fan, Chun-Hao Yang, and Baba~C Vemuri.
\newblock Horocycle decision boundaries for large margin classification in hyperbolic space.
\newblock \emph{arXiv preprint arXiv:2302.06807}, 2023.

\bibitem[Fang et~al.(2021)Fang, Harandi, and Petersson]{fang2021kernel}
Pengfei Fang, Mehrtash Harandi, and Lars Petersson.
\newblock Kernel methods in hyperbolic spaces.
\newblock In \emph{Proceedings of the IEEE/CVF International Conference on Computer Vision}, pp.\  10665--10674, 2021.

\bibitem[Fellbaum(2010)]{fellbaum2010princeton}
Christiane Fellbaum.
\newblock Princeton university: About wordnet, 2010.

\bibitem[Fern{\'a}ndez-Delgado et~al.(2014)Fern{\'a}ndez-Delgado, Cernadas, Barro, and Amorim]{fernandez2014we}
Manuel Fern{\'a}ndez-Delgado, Eva Cernadas, Sen{\'e}n Barro, and Dinani Amorim.
\newblock Do we need hundreds of classifiers to solve real world classification problems?
\newblock \emph{The journal of machine learning research}, 15\penalty0 (1):\penalty0 3133--3181, 2014.

\bibitem[Ganea et~al.(2018{\natexlab{a}})Ganea, B{\'e}cigneul, and Hofmann]{ganea2018hyperbolic}
Octavian Ganea, Gary B{\'e}cigneul, and Thomas Hofmann.
\newblock Hyperbolic entailment cones for learning hierarchical embeddings.
\newblock In \emph{International Conference on Machine Learning}, pp.\  1646--1655. PMLR, 2018{\natexlab{a}}.

\bibitem[Ganea et~al.(2018{\natexlab{b}})Ganea, B{\'e}cigneul, and Hofmann]{ganea2018nns}
Octavian Ganea, Gary B{\'e}cigneul, and Thomas Hofmann.
\newblock Hyperbolic neural networks.
\newblock \emph{Advances in neural information processing systems}, 31, 2018{\natexlab{b}}.

\bibitem[Garg et~al.(2022)Garg, Sani, and Anand]{garg2022learning}
Ashima Garg, Depanshu Sani, and Saket Anand.
\newblock Learning hierarchy aware features for reducing mistake severity.
\newblock In \emph{European Conference on Computer Vision}, pp.\  252--267. Springer, 2022.

\bibitem[Ghadimi~Atigh et~al.(2021)Ghadimi~Atigh, Keller-Ressel, and Mettes]{ghadimi2021hyperbolic}
Mina Ghadimi~Atigh, Martin Keller-Ressel, and Pascal Mettes.
\newblock Hyperbolic busemann learning with ideal prototypes.
\newblock \emph{Advances in Neural Information Processing Systems}, 34:\penalty0 103--115, 2021.

\bibitem[Ghadimi~Atigh et~al.(2022)Ghadimi~Atigh, Schoep, Acar, Van~Noord, and Mettes]{atigh2022hyperbolic}
Mina Ghadimi~Atigh, Julian Schoep, Erman Acar, Nanne Van~Noord, and Pascal Mettes.
\newblock Hyperbolic image segmentation.
\newblock In \emph{Proceedings of the IEEE/CVF conference on computer vision and pattern recognition}, 2022.

\bibitem[Girvan \& Newman(2002)Girvan and Newman]{girvan2002community}
Michelle Girvan and Mark~EJ Newman.
\newblock Community structure in social and biological networks.
\newblock \emph{Proceedings of the national academy of sciences}, 99\penalty0 (12):\penalty0 7821--7826, 2002.

\bibitem[Grinsztajn et~al.(2022)Grinsztajn, Oyallon, and Varoquaux]{grinsztajn2022tree}
L{\'e}o Grinsztajn, Edouard Oyallon, and Ga{\"e}l Varoquaux.
\newblock Why do tree-based models still outperform deep learning on typical tabular data?
\newblock \emph{Advances in Neural Information Processing Systems}, 35:\penalty0 507--520, 2022.

\bibitem[Heath et~al.(1993)Heath, Kasif, and Salzberg]{heath1993induction}
David Heath, Simon Kasif, and Steven Salzberg.
\newblock Induction of oblique decision trees.
\newblock In \emph{IJCAI}, volume 1993, pp.\  1002--1007. Citeseer, 1993.

\bibitem[Karthik et~al.(2021)Karthik, Prabhu, Dokania, and Gandhi]{karthik2021no}
Shyamgopal Karthik, Ameya Prabhu, Puneet~K Dokania, and Vineet Gandhi.
\newblock No cost likelihood manipulation at test time for making better mistakes in deep networks.
\newblock \emph{International Conference on Learning Representations}, 2021.

\bibitem[Kasarla et~al.(2022)Kasarla, Burghouts, van Spengler, van~der Pol, Cucchiara, and Mettes]{kasarla2022maximum}
Tejaswi Kasarla, Gertjan Burghouts, Max van Spengler, Elise van~der Pol, Rita Cucchiara, and Pascal Mettes.
\newblock Maximum class separation as inductive bias in one matrix.
\newblock \emph{Advances in Neural Information Processing Systems}, 35:\penalty0 19553--19566, 2022.

\bibitem[Katuwal \& Suganthan(2018)Katuwal and Suganthan]{katuwal2018enhancing}
Rakesh Katuwal and Ponnuthurai~N Suganthan.
\newblock Enhancing multi-class classification of random forest using random vector functional neural network and oblique decision surfaces.
\newblock In \emph{2018 International Joint Conference on Neural Networks}, pp.\  1--8. IEEE, 2018.

\bibitem[Katuwal et~al.(2020)Katuwal, Suganthan, and Zhang]{katuwal2020heterogeneous}
Rakesh Katuwal, Ponnuthurai~Nagaratnam Suganthan, and Le~Zhang.
\newblock Heterogeneous oblique random forest.
\newblock \emph{Pattern Recognition}, 99:\penalty0 107078, 2020.

\bibitem[Khrulkov et~al.(2020)Khrulkov, Mirvakhabova, Ustinova, Oseledets, and Lempitsky]{khrulkov2020hyperbolic}
Valentin Khrulkov, Leyla Mirvakhabova, Evgeniya Ustinova, Ivan Oseledets, and Victor Lempitsky.
\newblock Hyperbolic image embeddings.
\newblock In \emph{Proceedings of the IEEE/CVF Conference on Computer Vision and Pattern Recognition}, pp.\  6418--6428, 2020.

\bibitem[Klimovskaia et~al.(2020)Klimovskaia, Lopez-Paz, Bottou, and Nickel]{klimovskaia2020poincare}
Anna Klimovskaia, David Lopez-Paz, L{\'e}on Bottou, and Maximilian Nickel.
\newblock Poincar{\'e} maps for analyzing complex hierarchies in single-cell data.
\newblock \emph{Nature communications}, 11\penalty0 (1):\penalty0 2966, 2020.

\bibitem[Krizhevsky et~al.(2009)Krizhevsky, Hinton, et~al.]{krizhevsky2009learning}
Alex Krizhevsky, Geoffrey Hinton, et~al.
\newblock Learning multiple layers of features from tiny images.
\newblock 2009.

\bibitem[Liu et~al.(2019)Liu, Nickel, and Kiela]{liu2019hyperbolic}
Qi~Liu, Maximilian Nickel, and Douwe Kiela.
\newblock Hyperbolic graph neural networks.
\newblock \emph{Advances in neural information processing systems}, 2019.

\bibitem[Liu et~al.(2020)Liu, Chen, Pan, Ngo, Chua, and Jiang]{liu2020hyperbolic}
Shaoteng Liu, Jingjing Chen, Liangming Pan, Chong-Wah Ngo, Tat-Seng Chua, and Yu-Gang Jiang.
\newblock Hyperbolic visual embedding learning for zero-shot recognition.
\newblock In \emph{Proceedings of the IEEE/CVF conference on computer vision and pattern recognition}, 2020.

\bibitem[Lou et~al.(2020)Lou, Katsman, Jiang, Belongie, Lim, and De~Sa]{lou2020differentiating}
Aaron Lou, Isay Katsman, Qingxuan Jiang, Serge Belongie, Ser-Nam Lim, and Christopher De~Sa.
\newblock Differentiating through the fr{\'e}chet mean.
\newblock In \emph{International Conference on Machine Learning}, 2020.

\bibitem[Marconi et~al.(2020)Marconi, Ciliberto, and Rosasco]{marconi2020hyperbolic}
Gian Marconi, Carlo Ciliberto, and Lorenzo Rosasco.
\newblock Hyperbolic manifold regression.
\newblock In \emph{International Conference on Artificial Intelligence and Statistics}, pp.\  2570--2580. PMLR, 2020.

\bibitem[Menze et~al.(2011)Menze, Kelm, Splitthoff, Koethe, and Hamprecht]{menze2011oblique}
Bjoern~H Menze, B~Michael Kelm, Daniel~N Splitthoff, Ullrich Koethe, and Fred~A Hamprecht.
\newblock On oblique random forests.
\newblock In \emph{Machine Learning and Knowledge Discovery in Databases: European Conference, ECML PKDD 2011, Athens, Greece, September 5-9, 2011, Proceedings, Part II 22}, pp.\  453--469. Springer, 2011.

\bibitem[Mettes et~al.(2023)Mettes, Atigh, Keller-Ressel, Gu, and Yeung]{mettes2023hyperbolic}
Pascal Mettes, Mina~Ghadimi Atigh, Martin Keller-Ressel, Jeffrey Gu, and Serena Yeung.
\newblock Hyperbolic deep learning in computer vision: A survey.
\newblock \emph{arXiv preprint arXiv:2305.06611}, 2023.

\bibitem[Monath et~al.(2019)Monath, Zaheer, Silva, McCallum, and Ahmed]{monath2019gradient}
Nicholas Monath, Manzil Zaheer, Daniel Silva, Andrew McCallum, and Amr Ahmed.
\newblock Gradient-based hierarchical clustering using continuous representations of trees in hyperbolic space.
\newblock In \emph{Proceedings of the 25th ACM SIGKDD International Conference on Knowledge Discovery \& Data Mining}, pp.\  714--722, 2019.

\bibitem[Murthy et~al.(1993)Murthy, Kasif, Salzberg, and Beigel]{murthy1993oc1}
Sreerama~K Murthy, Simon Kasif, Steven Salzberg, and Richard Beigel.
\newblock Oc1: A randomized algorithm for building oblique decision trees.
\newblock In \emph{Proceedings of AAAI}, volume~93, pp.\  322--327. Citeseer, 1993.

\bibitem[Nickel \& Kiela(2017)Nickel and Kiela]{nickel2017poincare}
Maximillian Nickel and Douwe Kiela.
\newblock Poincar{\'e} embeddings for learning hierarchical representations.
\newblock In \emph{Advances in neural information processing systems}, volume~30, 2017.

\bibitem[Nickel \& Kiela(2018)Nickel and Kiela]{nickel2018learning}
Maximillian Nickel and Douwe Kiela.
\newblock Learning continuous hierarchies in the lorentz model of hyperbolic geometry.
\newblock In \emph{International conference on machine learning}, 2018.

\bibitem[Pan et~al.(2023)Pan, Chien, Tabaghi, Peng, and Milenkovic]{pan2023provably}
Chao Pan, Eli Chien, Puoya Tabaghi, Jianhao Peng, and Olgica Milenkovic.
\newblock Provably accurate and scalable linear classifiers in hyperbolic spaces.
\newblock \emph{Knowledge and Information Systems}, pp.\  1--34, 2023.

\bibitem[Peng et~al.(2021)Peng, Varanka, Mostafa, Shi, and Zhao]{peng2021hyperbolic}
Wei Peng, Tuomas Varanka, Abdelrahman Mostafa, Henglin Shi, and Guoying Zhao.
\newblock Hyperbolic deep neural networks: A survey.
\newblock \emph{IEEE Transactions on pattern analysis and machine intelligence}, 44\penalty0 (12):\penalty0 10023--10044, 2021.

\bibitem[Sebastian \& Sebastian(2023)Sebastian and Sebastian]{sebastian2023leveraging}
Rinu~Ann Sebastian and Anu~Maria Sebastian.
\newblock Leveraging semantic similarity to mitigate the severity of misclassification for safety critical applications.
\newblock \emph{Multimedia Tools and Applications}, 82\penalty0 (8):\penalty0 12615--12633, 2023.

\bibitem[Shimizu et~al.(2021)Shimizu, Mukuta, and Harada]{shimizu2021hyperbolic}
Ryohei Shimizu, Yusuke Mukuta, and Tatsuya Harada.
\newblock Hyperbolic neural networks++.
\newblock In \emph{International Conference on Learning Representations}, 2021.

\bibitem[Sonoda et~al.(2022)Sonoda, Ishikawa, and Ikeda]{sonoda2022fully}
Sho Sonoda, Isao Ishikawa, and Masahiro Ikeda.
\newblock Fully-connected network on noncompact symmetric space and ridgelet transform based on helgason-fourier analysis.
\newblock In \emph{International Conference on Machine Learning}, pp.\  20405--20422. PMLR, 2022.

\bibitem[Sun et~al.(2021{\natexlab{a}})Sun, Cheng, Zuberi, P{\'e}rez, and Volkovs]{sun2021hgcf}
Jianing Sun, Zhaoyue Cheng, Saba Zuberi, Felipe P{\'e}rez, and Maksims Volkovs.
\newblock Hgcf: Hyperbolic graph convolution networks for collaborative filtering.
\newblock In \emph{Proceedings of the Web Conference 2021}, 2021{\natexlab{a}}.

\bibitem[Sun et~al.(2021{\natexlab{b}})Sun, Zhang, Zhang, Wang, Peng, Su, and Philip]{sun2021hyperbolic}
Li~Sun, Zhongbao Zhang, Jiawei Zhang, Feiyang Wang, Hao Peng, Sen Su, and S~Yu Philip.
\newblock Hyperbolic variational graph neural network for modeling dynamic graphs.
\newblock In \emph{Proceedings of the AAAI Conference on Artificial Intelligence}, volume~35, pp.\  4375--4383, 2021{\natexlab{b}}.

\bibitem[Sur{\'\i}s et~al.(2021)Sur{\'\i}s, Liu, and Vondrick]{suris2021learning}
D{\'\i}dac Sur{\'\i}s, Ruoshi Liu, and Carl Vondrick.
\newblock Learning the predictability of the future.
\newblock In \emph{Proceedings of the IEEE/CVF Conference on Computer Vision and Pattern Recognition}, pp.\  12607--12617, 2021.

\bibitem[Tai et~al.(2022)Tai, Li, and Ku]{tai2022hyperbolic}
Chang-Yu Tai, Ming-Yao Li, and Lun-Wei Ku.
\newblock Hyperbolic disentangled representation for fine-grained aspect extraction.
\newblock In \emph{Proceedings of the AAAI Conference on Artificial Intelligence}, volume~36, pp.\  11358--11366, 2022.

\bibitem[Tifrea et~al.(2019)Tifrea, B{\'e}cigneul, and Ganea]{tifrea2018poincar}
Alexandru Tifrea, Gary B{\'e}cigneul, and Octavian-Eugen Ganea.
\newblock Poincar\'e glove: Hyperbolic word embeddings.
\newblock \emph{International Conference on Learning Representations}, 2019.

\bibitem[van Spengler et~al.(2023)van Spengler, Berkhout, and Mettes]{van2023poincar}
Max van Spengler, Erwin Berkhout, and Pascal Mettes.
\newblock Poincar\'e resnet.
\newblock In \emph{International Conference on Computer Vision}, 2023.

\bibitem[Weber et~al.(2020)Weber, Zaheer, Rawat, Menon, and Kumar]{weber2020robust}
Melanie Weber, Manzil Zaheer, Ankit~Singh Rawat, Aditya~K Menon, and Sanjiv Kumar.
\newblock Robust large-margin learning in hyperbolic space.
\newblock \emph{Advances in Neural Information Processing Systems}, 33:\penalty0 17863--17873, 2020.

\bibitem[Yang et~al.(2022)Yang, Zhou, Li, Liu, Pan, Xiong, and King]{yang2022hyperbolic}
Menglin Yang, Min Zhou, Zhihao Li, Jiahong Liu, Lujia Pan, Hui Xiong, and Irwin King.
\newblock Hyperbolic graph neural networks: a review of methods and applications.
\newblock \emph{arXiv preprint arXiv:2202.13852}, 2022.

\bibitem[Zachary(1977)]{zachary1977information}
Wayne~W Zachary.
\newblock An information flow model for conflict and fission in small groups.
\newblock \emph{Journal of anthropological research}, 33\penalty0 (4):\penalty0 452--473, 1977.

\end{thebibliography}
\bibliographystyle{tmlr_files/tmlr}

\clearpage

\begin{figure*}[!htbp]
  \centering
  \setlength\tabcolsep{2pt}
  \begin{tabular}{ccc}
   \includegraphics[width=0.3\linewidth]{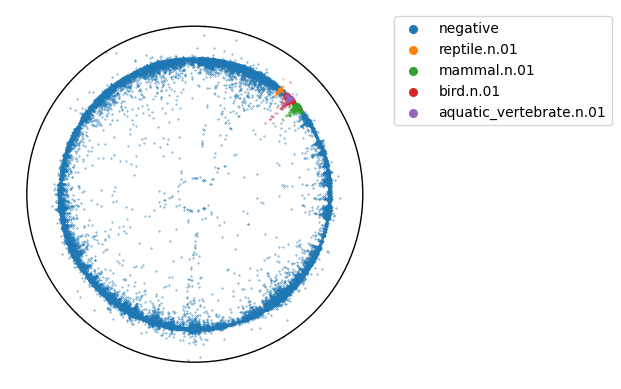} &
   \includegraphics[width=0.3\linewidth]{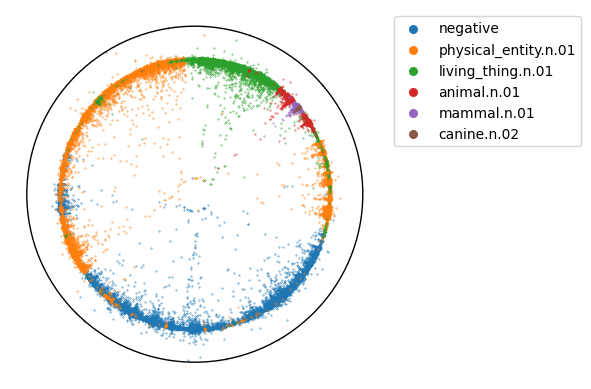} &
   \includegraphics[width=0.3\linewidth]{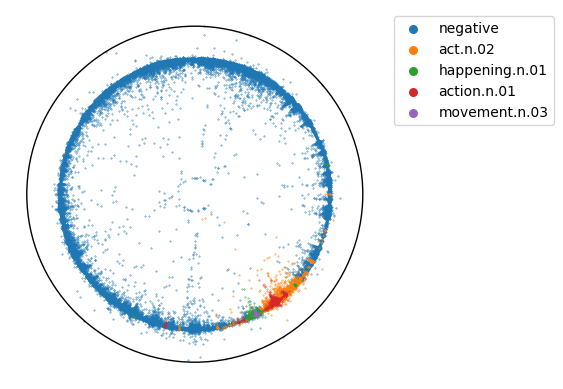} \\
   (a) & (b) & (c) \\
  \end{tabular}
    \caption{\textbf{Visualizing the multi-class WordNet experiments.} We embed all WordNet nouns into two dimensions with hyperbolic entailment cones~\cite{ganea2018hyperbolic} and color the different classes of our three multi-class WordNet experiments. } 
    \label{fig:vis}
\end{figure*}

\section{Supplementary Material}

\subsection{Dataset details}

\paragraph{Multi-class WordNet}

For the first experiment, we take four subtrees of the \texttt{vertebrate.n.01} subtree: \texttt{reptile.n.01} (237 samples), \texttt{mammal.n.01} (944 samples), \texttt{bird.n.01} (697 samples), and \texttt{aquatic\_vertebrate.n.01} (506 samples). Together with the negative class of 63307 samples, this leads to a 5-way classification problem. 

For the second experiment, we take nested subtrees: \texttt{physical\_entity.n.01} (21252 samples), \texttt{living\_thing.n.01} (12460 samples), \texttt{animal.n.01} (2268 samples), \texttt{mammal.n.01} (767 samples), and \texttt{canine.n.01} (178 samples). Together with the negative class of 28766 samples, this leads to a 6-way classification problem. 

For the third experiment, we combine the two. We take the \texttt{act.n.01} subtree (4182 samples), as well as its subtree \texttt{happening.n.01} (639 samples). Additionally, we include \texttt{action.n.01} (1317 samples) and its subtree \texttt{movement.n.03} (141 samples). Together with the negative class of 59412 samples, this leads to a 5-way classification problem. 

Visualizations in 2D for the three experiments are given in Fig.~\ref{fig:vis}. Due to the multi-class setting and large class imbalance in all experiments, we use the macro-f1 to evaluate methods.

\paragraph{WordNet}

The 82115 nodes of the WordNet noun hierarchy are split into two classes for each experiment. The corresponding number of samples in the train and test split for the negative and positive classes are given in Tab.~\ref{tab:wn_details}. Due to the large class imbalance in all experiments, we use the AUPR to evaluate methods.

\begin{table*}[h]
\centering
% \resizebox{\linewidth}{!}{
\begin{tabular}{lcccccccc}
\toprule
& \multirow{2}{*}{Animal} & \multirow{2}{*}{Group} & \multirow{2}{*}{Worker} & \multirow{2}{*}{Mammal} & \multirow{2}{*}{Tree} & \multirow{2}{*}{Solid} & \multirow{2}{*}{Occupation}  & \multirow{2}{*}{Rodent}\\
 &&&&&&&\\
\midrule
\rowcolor{mygray} \textbf{Train} & & & & & & & & \\
Positive & 3213 & 6701 & 892 & 945 & 811 & 985 & 226 & 110 \\
Negative & 62478 & 58990 & 64799 & 64746 & 64880 & 64706 & 65465 & 65581\\
\midrule
\rowcolor{mygray} \textbf{Test} & & & & & & & &\\
Positive & 803 & 1675 & 223 & 236 & 203 & 246 & 57 & 28\\
Negative & 15620 & 14748 & 16200 & 16187 & 16220 & 16177 & 16366 & 16395 \\
\bottomrule
\end{tabular}
% }
\caption{\textbf{WordNet dataset details.} We show the class distribution for the train and test split per experiment.}
\label{tab:wn_details}
\end{table*}

\paragraph{Networks}

As described in the main text, the networks are (1) \karate~\citep{zachary1977information}, with 34 nodes and 2 classes, (2) \polblogs~\citep{adamic2005political}, with 1224 nodes and 2 classes, (3) football~\citep{girvan2002community}, with 115 nodes and 12 classes, and (4) \polbooks\footnote{http://www-personal.umich.edu/~mejn/netdata/}, with 105 nodes and 3 classes.

\subsection{Full implementation details}

We run a grid search for all methods. For the linear SVM, we search over $C \in \{-5, -4, \cdots, 15\}$. For the RBF SVM, we search over $C \in \{-5, -1, \cdots, 15\}$ and $\gamma \in \{-15, -11, \cdots, 1, 3\}$. For the random forest and oblique random forest, we search over the hyperparameter that controls at what number of samples to stop splitting a node $m \in \{1, 3, 5\}$ on the small network datasets and $m \in \{3, 7, 11\}$ on the larger WordNet datasets. For the hyperbolic logistic regression, we search over the learning rate $lr \in \{0.1, 0.01, 0.001\}$, and the batch size $bs \in \{16, 32, 64\}$ on the network datasets and $bs \in \{128, 256, 512\}$ on the WordNet datasets. For HoroSVM, we search over $C \in \{-5, -4, \cdots, 15\}$. For HoroRF, we search over $\beta \in \{0, 0.9, \cdots, 0.9999\}$, and $m \in \{1, 3, 5\}$ on the network datasets and $m \in \{3, 7, 11\}$ on the WordNet datasets.

We run all our experiments on a machine running CentOS 7.9.2009 with an AMD EPYC 7302 16-Core Processor with access to 48GB of memory. We set our seeds such that our implementation of the HoroSVM baseline resembles the results on the network datasets reported in its original publication. 

\subsection{Tabular data}

We aim to investigate how our method works on real-world tabular data. We compare HoroRF against the Euclidean classifiers on experiments 1-3 of the UCI121 benchmark (acute-nephritis, oocytes-merluccius-states-2f, and primary-tumor)~\cite{fernandez2014we}, using the preprocessed versions of~\citet{cai2021arm}. We find that mapping these Euclidean datasets to the Poincar\'e ball with a curvature of $1$ leads to all samples lying on the boundary. For this reason, we also search over the curvature parameter in $c\in[1, 0.1, 0.01]$. We show results on the datasets in Tab.~\ref{tab:tabular}, finding that HoroRF is competitive with the Euclidean classifiers despite being designed for a different data representation. 

\begin{table*}[t]
\centering
% \resizebox{\linewidth}{!}{
\begin{tabular}{lccc}
\toprule
& \multirow{2}{*}{\an} & \multirow{2}{*}{\oms} & \multirow{2}{*}{\pt}\\
 &&&\\
\midrule
\rowcolor{mygray} \textbf{Euclidean} & & &\\
LinSVM & 100\small{$\pm0.0$} & 91.4\small{$\pm1.1$} & 41.2\small{$\pm0.0$}\\
RBFSVM & 100\small{$\pm0.0$} & 92.8\small{$\pm0.0$} & 43.0\small{$\pm0.0$}\\
RF & 97.8\small{$\pm1.6$} & 91.3\small{$\pm0.2$} & 47.7\small{$\pm0.8$}\\
OblRF & 100\small{$\pm0.0$} & 92.3\small{$\pm0.2$} & 42.4\small{$\pm0.9$}\\
\midrule
\rowcolor{mygray} \textbf{Hyperbolic} & & &\\
\emph{HoroRF} & 100\small{$\pm0.0$} & 90.5\small{$\pm0.4$} & 44.0\small{$\pm1.1$}\\
\bottomrule
\end{tabular}
% }
\caption{\textbf{Comparative evaluation on tabular experiments.} We run a grid search to determine the optimal configuration on the validation set and report the mean and standard deviation of the micro f1-score on the test set averaged over three runs. HoroRF obtains competitive performance.}
\label{tab:tabular}
\end{table*}

\subsection{Different embeddings}

\paragraph{Euclidean embeddings}

All results in the main paper are obtained by fitting the methods to the original hyperbolic embeddings. In Tabs.~\ref{tab:wordnet_e}-~\ref{tab:networks_e}, we show that applying the Euclidean methods to the original hyperbolic embeddings outperforms applying them to the embeddings mapped to Euclidean space.

\begin{table*}[t]
\centering
\resizebox{\linewidth}{!}{
\begin{tabular}{lcccccccc}
\toprule
& \multirow{2}{*}{Animal} & \multirow{2}{*}{Group} & \multirow{2}{*}{Worker} & \multirow{2}{*}{Mammal} & \multirow{2}{*}{Tree} & \multirow{2}{*}{Solid} & \multirow{2}{*}{Occupation}  & \multirow{2}{*}{Rodent}\\
 &&&&&&&\\
\midrule
\rowcolor{mygray} \textbf{Euc. embeddings} & & & & & & & & \\
LinSVM & 2.5\small{$\pm0.0$} & 5.6\small{$\pm0.1$} & 0.7\small{$\pm0.0$} & 0.7\small{$\pm0.0$} & 0.6\small{$\pm0.0$} & 0.8\small{$\pm0.0$} & 0.2\small{$\pm0.0$} & 0.1\small{$\pm0.0$} \\
RBFSVM & 2.5\small{$\pm0.0$} & 5.4\small{$\pm0.0$} & 0.7\small{$\pm0.0$} & 0.7\small{$\pm0.0$} & 0.6\small{$\pm0.0$} & 0.8\small{$\pm0.0$} & 0.2\small{$\pm0.0$} & 0.1\small{$\pm0.0$} \\
RF & 98.5\small{$\pm$0.2} & 96.4\small{$\pm$0.1} & 72.5\small{$\pm$1.0} & 98.4\small{$\pm$1.0} & \textbf{77.2}\small{$\pm$2.3} & 91.2\small{$\pm$0.8} & 63.4\small{$\pm$0.6} & 28.6\small{$\pm$6.5}\\
OblRF & 98.1\small{$\pm$0.1} & 96.2\small{$\pm$0.2} & 71.4\small{$\pm$0.8} & 98.8\small{$\pm$0.7} & 77.0\small{$\pm$1.6} & 91.4\small{$\pm$0.7} & 62.9\small{$\pm$0.3} & 29.3\small{$\pm$7.4} \\
\midrule
\rowcolor{mygray} \textbf{Hyp. embeddings} & & & & & & & & \\
LinSVM & 2.5\small{$\pm0.0$} & 6.0\small{$\pm0.6$} & 0.7\small{$\pm0.0$} & 0.7\small{$\pm0.0$} & 0.6\small{$\pm0.0$} & 0.8\small{$\pm0.0$} & 0.2\small{$\pm0.0$} & 0.1\small{$\pm0.0$} \\
RBFSVM & 2.5\small{$\pm0.0$} & 6.0\small{$\pm0.6$} & 0.7\small{$\pm0.0$} & 0.7\small{$\pm0.0$} & 0.6\small{$\pm0.0$} & 0.8\small{$\pm0.0$} & 0.2\small{$\pm0.0$} & 0.1\small{$\pm0.0$} \\
RF & \textbf{98.6}\small{$\pm$0.2} &  96.6\small{$\pm$0.4} & 70.0\small{$\pm3.1$} & \textbf{99.3}\small{$\pm$0.3} & 75.8\small{$\pm2.0$} & 92.5\small{$\pm$ 1.2} & 59.8\small{$\pm1.1$} & 34.5\small{$\pm3.0$}\\
OblRF & 98.4\small{$\pm$0.1} & \textbf{96.7} \small{$\pm$0.3} & \textbf{73.3}\small{$\pm$1.3} & 98.8\small{$\pm$0.3} & 76.2\small{$\pm$3.2} & \textbf{92.9}\small{$\pm$1.7} & \textbf{65.0}\small{$\pm$2.4} & \textbf{38.2}\small{$\pm$0.9}\\
\bottomrule
\end{tabular}
}
\caption{\textbf{Comparing Euclidean methods applied to different embeddings on binary WordNet experiments.} We follow the experimental protocol of~\cite{fan2023horocycle}. \textbf{Bold} denotes the best method. Euclidean classifiers perform better on the hyperbolic embeddings.}
\label{tab:wordnet_e}
\end{table*}

\begin{table}[t!]
\centering
% \resizebox{\linewidth}{!}{
\begin{tabular}{lccc}
\toprule
& Same & \multirow{2}{*}{Nested} & \multirow{2}{*}{Both} \\
& level\\
\midrule
\rowcolor{mygray} \textbf{Euc. embeddings} & & &\\
LinSVM & 47.7\small{$\pm$1.3} & 51.0\small{$\pm$0.3} & 32.9\small{$\pm$10.2} \\
RBFSVM & 81.3\small{$\pm$2.1} & 89.8\small{$\pm$1.0} & 73.8\small{$\pm$0.6} \\
RF & 88.9\small{$\pm$1.0} & 92.0\small{$\pm$0.4} & 81.1\small{$\pm$0.4}\\
OblRF & 90.5\small{$\pm$0.7} & 92.2\small{$\pm$0.2} & \textbf{82.0}\small{$\pm$0.7}\\
\midrule
\rowcolor{mygray} \textbf{Hyp. embeddings} & & &\\
LinSVM & 48.8\small{$\pm$0.9} & 59.7\small{$\pm$0.5} & 35.6\small{$\pm$0.2} \\
RBFSVM & 80.0\small{$\pm$1.8} & 89.8\small{$\pm$1.1} & 70.8\small{$\pm$0.6} \\
RF & 89.7\small{$\pm$0.7} & 91.7\small{$\pm$0.3} & 81.5\small{$\pm$1.5} \\
OblRF &  \textbf{91.3}\small{$\pm$0.6} & \textbf{93.0}\small{$\pm$0.2} & 81.3\small{$\pm$1.4}\\
\bottomrule
\end{tabular}
% }
\caption{\textbf{Comparing Euclidean methods applied to different embeddings on multi-class WordNet experiments.} We follow our experimental protocol as described in the main paper. \textbf{Bold} denotes the best method. Euclidean classifiers perform better on the hyperbolic embeddings.}
\label{tab:wordnet_multi_e}
\end{table}

\begin{table}[t!]
% \resizebox{\linewidth}{!}{
\centering
\begin{tabular}{lcccc}
\toprule
& \multicolumn{2}{c}{Binary} & \multicolumn{2}{c}{Multi-class}\\
\cmidrule(lr){2-3} \cmidrule(lr){4-5}
& Karate & Polbooks & Football & Polblogs\\
\midrule
\rowcolor{mygray} \textbf{Euc. embeddings} & & & & \\
LinSVM & \textbf{95.4}\small{$\pm$2.3} & 92.4\small{$\pm$0.3} & 33.4\small{$\pm$5.2} & 85.7\small{$\pm$0.6}\\
RBFSVM & \textbf{95.4}\small{$\pm$2.3} & \textbf{92.5}\small{$\pm$0.2} & 35.1\small{$\pm$3.3} & 83.6\small{$\pm$1.5}\\
RF & 94.3\small{$\pm$3.1} & 92.1\small{$\pm$0.3} & 35.0\small{$\pm$5.1} & 84.8\small{$\pm$1.9}\\
OblRF & 94.9\small{$\pm$3.3} & 92.0\small{$\pm$0.6} & 35.3\small{$\pm$2.5} & 83.8\small{$\pm$1.2} \\
\midrule
\rowcolor{mygray} \textbf{Hyp. embeddings} & & & & \\
LinSVM & \textbf{95.4}\small{$\pm$2.3} & 92.4\small{$\pm$0.3} & 33.2\small{$\pm$5.1} & 85.5\small{$\pm$0.9}\\
RBFSVM & \textbf{95.4}\small{$\pm$2.3} & 92.4\small{$\pm$0.3} & 35.5\small{$\pm$4.7} & 84.4\small{$\pm$1.9}\\
RF & 94.3\small{$\pm$3.1} & 92.1\small{$\pm$0.3} & 36.2\small{$\pm$4.9} & 85.1\small{$\pm$2.1}\\
OblRF & 94.8\small{$\pm$2.2} & 92.1\small{$\pm$0.3} & \textbf{36.7}\small{$\pm$2.7} & 84.4\small{$\pm$1.3}\\
\bottomrule
\end{tabular}
% }
\caption{\textbf{Comparing Euclidean methods applied to different embeddings on network datasets.} We follow the experimental protocol of~\cite{fan2023horocycle}. \textbf{Bold} denotes the best method. Results are similar between embeddings.}
\label{tab:networks_e}
\end{table}

\paragraph{Hyperboloid embeddings \& comparison to HyperRF}

The concurrent work of \citet{chlenski2023fast} finds splits as midpoints between angles of data points in the hyperboloid model of hyperbolic space, rather than the Poincar\'e ball model used in our work. We compare HoroRF to HyperRF on the multi-class networks, the four most imbalanced WordNet experiments, and the same-level and nested multi-class experiments in Tab.~\ref{tab:hyprf}. We find that HoroRF outperforms HyperRF on the networks, they perform similarly on the WordNet subtree classification, and HyperRF slightly outperforms HoroRF on the multi-class WordNet experiments.

% However, comparing HyperRF to HoroRF involves not only different algorithms but also different hyperbolic models. Given this, there is no reason to fix the Poincar\'e ball curvature to one. For instance, HoroRF with a curvature of $0.1$ and $0.01$ on the \emph{Same level} experiments reach a performance of 91.8\small{$\pm$0.3} and 92.3\small{$\pm$0.3}, respectively, showing that in this representation space, HoroRF comes out on top. Further research is needed into the best embedding model for hyperbolic classification.

\begin{table}[t!]
\resizebox{\linewidth}{!}{
\centering
\begin{tabular}{lccccccccc}
\toprule
& \multirow{2}{*}{Football} & \multirow{2}{*}{Polblogs} & \multirow{2}{*}{Worker} & \multirow{2}{*}{Tree} & \multirow{2}{*}{Occupation}  & \multirow{2}{*}{Rodent} & Same & \multirow{2}{*}{Nested} & \multirow{2}{*}{Mean}\\
& &&&&&&level &\\
\midrule
HyperRF & 33.0\small{$\pm$2.3} & 84.4\small{$\pm$1.8} & 71.5\small{$\pm$ 0.9} &  \textbf{77.5}\small{$\pm$ 2.1} & 53.9\small{$\pm$ 4.4} & \textbf{40.3}\small{$\pm$ 1.6} & \textbf{91.7}\small{$\pm$1.0} & \textbf{93.8} \small{$\pm$0.8} & 68.3 \\
\emph{HoroRF} & \textbf{38.3}\small{$\pm$1.8} & \textbf{86.1}\small{$\pm$1.0} & \textbf{73.4}\small{$\pm$ 1.7} &  76.6\small{$\pm$ 3.8} & \textbf{65.6}\small{$\pm$ 1.5} & 39.0\small{$\pm$ 2.4} & 91.3\small{$\pm$0.3} & 93.3\small{$\pm$1.1} & \textbf{70.4}\\
\bottomrule
\end{tabular}
}
\caption{\textbf{Comparative evaluation between HyperRF and HoroRF.} We follow the experimental protocol for each experiment as described in the main text. \textbf{Bold} denotes the best method. The methods are comparable across the benchmarks.}
\label{tab:hyprf}
\end{table}

% \clearpage

% \input{rebuttal}

%\appendix
%\section{Appendix}
%You may include other additional sections here.

\end{document}